%% file: main+supp.tex
\documentclass[11pt,a4paper,dvipsnames]{article}
\usepackage[hyperref]{arxiv}
\usepackage{times}
\usepackage{latexsym}

\usepackage{microtype}

\finalcopy

\usepackage{epsfig}
\usepackage{graphicx}
\usepackage{amsmath}
\usepackage{amssymb}

\usepackage{url}            % simple URL typesetting

\usepackage{booktabs}       % professional-quality tables
\usepackage{amsfonts}       % blackboard math symbols
\usepackage{mathtools}
\usepackage{amssymb}
\usepackage{nicefrac}       % compact symbols for 1/2, etc.
\usepackage{paralist}
\usepackage{caption}
\usepackage{subcaption}
\usepackage{multirow}
\usepackage[acronym,smallcaps,nowarn,section,nogroupskip,nonumberlist]{glossaries}
\usepackage{siunitx}
\usepackage{tikz}
\usepackage{xr}
\usepackage[nameinlink,capitalise]{cleveref}
\usepackage{titlesec}
\usepackage{setspace}

\definecolor{red}{HTML}{E41A1C}
\definecolor{orange}{HTML}{FF7F00}
\definecolor{yellow}{HTML}{FFC020}
\definecolor{green}{HTML}{4DAF4A}
\definecolor{blue}{HTML}{377EB8}
\definecolor{purple}{HTML}{984EA3}
\usetikzlibrary{calc,backgrounds,positioning,bayesnet}
\graphicspath{{images/}}
\crefname{figure}{Figure}{Figures}
\crefname{table}{Table}{Tables}
\crefname{subtable}{Table}{Tables}
\crefname{section}{\S}{\S\S}
\Crefname{section}{\S}{\S\S}
\crefformat{equation}{Eq.~(#2#1#3)}
\crefformat{enumi}{(#2#1#3)}
\glsdisablehyper{}
\newacronym{AMT}{amt}{Amazon Mechanical Turk}
\newacronym{MR}{mr}{mean rank}
\newacronym{MRR}{mrr}{mean reciprocal rank}
\newacronym{CCA}{cca}{canonical correlation analysis}
\newacronym{VQA}{vqa}{visual question answering}
\newacronym{VD}{vd}{visual dialogue}
\newacronym{SGD}{sgd}{stochastic gradient descent}
\newacronym{SOTA}{sota}{state-of-the-art}
\newacronym{NDCG}{ndcg}{normalised discounted cumulative gain}
\newacronym{DCG}{dcg}{discounted cumulative gain}
\newacronym{IOU}{iou}{intersection-over-union}
\newacronym{ML}{ml}{machine learning}
\titlespacing*{\section}
{0pt}{1ex plus 0.5ex minus .2ex}{1ex plus .2ex}
\titlespacing*{\subsection}
{0pt}{0.8ex plus 0.5ex minus .2ex}{0.8ex plus .2ex}

\setdefaultleftmargin{2ex}{2ex}{2ex}{}{}{}

\newcommand{\newdataset}{\textit{DenseVisDial}}
\newcommand{\bfx}{\mathbf{x}}
\newcommand{\gtA}{A_{\text{gt}}}
\newcommand{\norm}[1]{\left\lVert#1\right\rVert}
\input{maths-basic}             % import extra maths

\newcommand{\rank}[2][black]{\textcolor{#1}{\textcircled{\scalebox{0.6}{#2}}}}

\begin{document}

%%%%%%%%% TITLE
\title{A Revised Generative Evaluation of Visual Dialogue}

\author{%
  Daniela Massiceti \quad
  Viveka Kulharia \quad
  Puneet K. Dokania \quad
  N. Siddharth \quad
  Philip H.S. Torr\\
  University of Oxford\\
  {\tt\small \{daniela, viveka, puneet, nsid, phst\} @robots.ox.ac.uk}
}

\maketitle
%\thispagestyle{empty}

%%%%%%%%% ABSTRACT
\begin{abstract}
  Evaluating \gls{VD}, the task of answering a sequence of questions relating to a visual input, remains an open research challenge.
  The current evaluation scheme of the \textit{VisDial} dataset computes the ranks of ground-truth answers in predefined candidate sets, which~\citet{massiceti2018novd} show can be susceptible to the exploitation of dataset biases.
  This scheme also does little to account for the different ways of expressing the same answer---an aspect of language that has been well studied in \textsc{nlp}.
  We propose a revised evaluation scheme for the \textit{VisDial} dataset leveraging metrics from the \textsc{nlp} literature to measure consensus between answers \emph{generated} by the model and a \emph{set} of relevant answers.
  We construct these relevant answer sets using a simple and effective semi-supervised method based on correlation, which allows us to automatically extend and scale sparse relevance annotations from humans to the entire dataset.
  We release these sets and code for the revised evaluation scheme as \newdataset\footnote{\url{https://github.com/danielamassiceti/geneval_visdial}} and intend them to be an improvement to the dataset in the face of its existing constraints and design choices.
\end{abstract}

\glsresetall
\section{Introduction}

\label{sec:intro}
\begin{figure}[t!]
  \centering\scriptsize
  \vspace*{-2ex}
  \begin{tabular}{@{}c@{}}
    \includegraphics[width=0.65\linewidth]{hook.jpg} \\
    \begin{tabular*}{0.72\linewidth}{@{}l@{\extracolsep{\fill}}r@{}}
      \textbf{Question}               & \textbf{Answer} \\
      \cmidrule{1-1}                  \cmidrule{2-2}
      How old is the baby?            & About 2 years old \\
      What color is the remote?       & White \\
      Where is the train?             & On the road \\
      How many cows are there?        & Three \\
    \end{tabular*}
  \end{tabular}
  \vspace*{-2ex}
  \caption{%
    Failures in \emph{visual} dialogue (from~\citet{massiceti2018novd})---answers are unrelated to the image. Biases in the \textit{VisDial} dataset, compounded by a rank-based evaluation, can mislead progress on the \acrshort{VD} task.
    %\protect\footnotemark
  }
  \label{fig:another-vd-fail}
  \vspace*{-1ex}
\end{figure}
%\footnotetext{From online demos of \acrshort{SOTA} models--\href{http://demo.visualdialog.org/}{VisDial}~\citep{Das_VisualDialog} and \href{http://www.robots.ox.ac.uk/~daniela/research/flipdial/1vd_demo}{FlipDial}~\citep{massiceti2018}}

The growing interest in visual conversational agents~\citep{Antol_VQA,Das_VisualDialog,de2017guesswhat,johnson2017clevr} has motivated the need for automated evaluation metrics for the responses generated by these agents.
Robust evaluation schemes, however, are an open research challenge~\citep{mellish1998evaluation,reiter2009investigation}.
This is the case for \textit{VisDial}~\citep{Das_VisualDialog}, a dataset targeting the \gls{VD} task---answering a sequence of questions about an image given a history of previous questions and answers.
At test time, a trained model is used to score fixed sets of candidate answers for each test question, and a suite of rank-based metrics are computed on the ranked sets: single-candidate metrics which are a function of the ground-truth (\textsc{gt}) answer's position, and a multi-candidate metric which weighs the ranked set with relevance scores assigned from human annotation.
A limit of this scheme is that a simple model~\citep{massiceti2018novd} based on \gls{CCA}, which learns to maximise correlation between questions and answers while completely ignoring the image and dialogue history, is comparable in \gls{MR}---one of the dataset's primary rank-based metrics---to \gls{SOTA} models, all of which employ complex neural network architectures of millions of parameters, requiring many hours of \textsc{gpu}(s) training.
%
%In comparison, \gls{CCA} uses standard off-the-shelf feature extractors, avoids computing gradients, involves few parameters and requires just a few seconds on {\sc cpu}---all without the image or dialogue!
This suggests that exploitable biases exist within the \textit{VisDial} dataset, whose effects are compounded by a rank-based evaluation ill-suited to the \gls{VD} task.

Motivated by this, we propose a revised, more robust, evaluation scheme for the \gls{VD} task, informed by the key shortfalls of the current evaluation, namely that
\begin{inparaenum}[1)]
\item the candidate sets contain multiple, \emph{equally feasible} answers, rendering both single- and multi-candidate ranking metrics less meaningful, and
\item the evaluation is an indirect ranking task, rather than a direct assessment of the answers \emph{generated} by a model---the goal of a \gls{VD} agent.
\end{inparaenum}

% \vspace*{-0.2em}
Our revised evaluation instead adopts standard metrics from \textsc{nlp} to measure the similarity between an answer \emph{generated} by a model and a reference set of answers for the image and question.
This aligns better with the generative nature of a true \gls{VD} agent, and with established evaluation set-ups for other language generations tasks, including \acrshort{VQA}~\citep{Antol_VQA} and image captioning~\citep{ChenFLVGDZ15,young2014image}.
%evaluation set-up which compares predicted answers to answers provided by \emph{multiple} human annotators.
%
Unlike the current evaluation, it also accounts for diversity in answer generations, which we compare across models.

For \textit{VisDial}, however, the answer relevance scores used to construct the reference sets are only available for a small subset of the dataset.
Drawing on the pseudo-labelling literature for semi-supervised learning~\citep{lee2013pseudo,wu2006fuzzy,iscen2019label}, we develop a semi-supervised approach which leverages these human annotations, and automatically learns to extract the relevant answers from a given candidate set.
We apply \gls{CCA} between pre-trained question and answer features, channelling~\citet{massiceti2018novd}, and use a clustering heuristic on the resulting correlations to extract the candidate answers most correlated with the ground-truth answer---the reference sets, or pseudo-labels.
This was inspired by prior work showing the surprising strength of simple, non-deep baselines~\citep{ZhouTSSF15,massiceti2018novd,strang2018don,anand2018blindfold}.
Using this approach, we automatically construct sets of relevant answers for the entire \textit{VisDial} dataset, which we validate in multiple ways, including with human judgements via \gls{AMT}, and release as \newdataset{}---a dense version of \textit{VisDial}.
Using this data and the revised scheme, we re-evaluate existing \gls{SOTA} models and gain new insights into model differences, specifically generation diversity, otherwise unavailable with the existing evaluation.
The scheme also improves on the existing multi-candidate ranking evaluation, only applicable for \(1/10\) of the dataset due to the cost and time of collecting relevance scores from humans.
Finally, while we use these reference sets exclusively for better evaluation purposes, we also show that using them as (for-free) additional training data can improve performance, which is promising for future progress on the \gls{VD} task.
%

%These metrics are more cleanly able to separate performance between baseline and \gls{SOTA} models, and importantly depart from the issues of the current rank-based approach.
%
%ssessing the quality and relevance of the generated answers is equally, if not more, challenging, given the additional constraints of the conditioning image.
%
%While a simple task for a human evaluator, automatic evaluation metrics, which have been a large focus in the community to assess at scale, often struggle to capture the nuanced differences in quality of natural language generations, in particular for down-stream tasks beyond machine translation.
%
%While some metrics have been shown to be closely correlated with human perception, for example {\sc meteor}~\citep{Papineni_2002bleu} and {\sc cider}~\citep{Vedantam_2015cider},

%% Sid & Puneet: Do we really need this?
% We consider \newdataset{} and the revised evaluation to be an improvement of the \textit{VisDial} dataset, under the constraints of its existing data biases.
% %
% Going forwards, evaluation metrics should look beyond syntactic-based matching schemes, and toward metrics where performance on a downstream task, necessitating use of language, is measured~\citep{de2017guesswhat,paul2017visual,visdial_rl,elasri2017frames}.
% %
% More generally, however, we acknowledge that the evaluation of language is an open research challenge, and we recognise that all efforts, including this work, are with the intention to move the field forwards.

\noindent%
To summarise, our contributions are:
\begin{compactenum}
\item A revised evaluation scheme for the \textit{VisDial} dataset based on metrics from \textsc{nlp} which measure the similarity between a \emph{generated} answer and a reference set of feasible answers.
\item A semi-supervised method for automatically extracting these reference sets from given candidate sets at scale, verified using human annotators.
% \item A semi-supervised method for automatically extracting these reference sets from given candidate sets for questions and images which we verify using human annotators.
  %
\item An expanded \newdataset{} data with the automatically constructed reference sets, released for future evaluation and model development.
\end{compactenum}

\section{Preliminaries}
\label{sec:prelim}

We first define generative \gls{VD} (for \textit{VisDial}), and how~\citet{massiceti2018novd} employ \gls{CCA} for this.

\subsection{Visual Dialogue (\acrshort{VD}) \& \textit{VisDial} dataset}
\label{sec:prelim-vd}

Given image~\(I\) and dialogue history \([(Q_1, A_1), (Q_2, A_2), \ldots, Q_i]\), the generative \gls{VD} task involves generating answer \(A_i\).

The principal approach towards \gls{VD} has been facilitated by the \textit{VisDial} dataset~\citep{Das_VisualDialog}, a large corpus of images paired with question-answer dialogues, sequentially collected by pairs of annotators in an interactive game on \gls{AMT}.
{\it VisDial v1.0} comprises \(123,287/2064/8000\) train/val/test images, each paired with dialogues of up to 10 exchanges\footnote{10 exchanges for train/val, and \(\leq\) 10 exchanges for test.}.
Each question is coupled with a candidate set of 100 answers~\(\bm{A}\) including a ground-truth answer~\(\gtA \in \bm{A}\).
For a subset (2000/2064/8000), one question per image contains human-annotated relevance scores~\(\rho(A) \in [0,1] \, \text{where\,} A \in \bm{A}\).

The generative \gls{VD} task learns to \emph{generate} answers conditioned on image-question pairs using only \((I,Q,\gtA)\) triplets~\citep{Das_VisualDialog}.
At test time, given an image-question pair, each answer in its associated candidate answers \(\bm{A}\) is scored under the model's learned likelihood.
The rank of~\(\gtA\) is then used to judge the model's effectiveness at the \gls{VD} task, averaged over the dataset to get the \gls{MR}.
Other metrics also computed include \gls{NDCG} on the candidate answers' human-annotated relevance scores.

In a second paradigm introduced by~\citet{Das_VisualDialog}, the model instead uses the full \((I,Q,\bm{A})\) at train time, and simply frames the predictive task as a classification problem of selecting~\(\gtA\) out of~\(\bm{A}\).
At test time, the candidates are then directly scored by the classifier's \textit{softmax} probabilities.
We argue that this discriminative setting is an over-simplification of the \gls{VD} task: answering questions is not simply selecting the correct answer from a set.
The focus of the remainder of this paper is therefore fully on the generative \gls{VD} task.

\glsreset{MR}
\subsection{Canonical Correlation Analysis for \gls{VD}}
\label{sec:ccaforvd}

\gls{CCA}, applied between question and answers, achieves near-\gls{SOTA} \gls{MR} on the \textit{VisDial} dataset~\citep{massiceti2018novd}.
%
%Surprisingly, this approach ignores the dialogue history and image completely, and takes a fraction of the training time and parameters of complex \gls{SOTA} neural networks.
%
Inspired by this result and the extreme simplicity of \gls{CCA}, we introduce this formulation with reference to \gls{VD}.
%
%We note that \gls{CCA} has the benefit of:
%
%\begin{inparaenum}[i)]
%\item explicitly targeting correlations, which is ideally suited to examining the relationship between questions and answers (and images) for the \gls{VD} task, and
%\item being extremely simple, avoiding complex models, gradient computation, and large {\sc gpu} resources and time.
%\end{inparaenum}
%
%Inspired by this result, we adopt \gls{CCA} as a diagnostic tool to probe \textit{VisDial}'s current evaluation paradigm.

% \vspace*{-0.5em}
% \paragraph{\gls{CCA}:} \label{sec:prelim-cca}
Given paired observations $\{\bfx_1 \in \mathbb{R}^{n_1}, \bfx_2 \in \mathbb{R}^{n_2}\}$, \gls{CCA} jointly learns projections \( W_1 \in \mathbb{R}^{n_1 \times k} \) and \( W_2 \in \mathbb{R}^{n_2 \times k} \), \( k \leq \min (n_1,n_2) \), which are maximally correlated~\citep{hotelling1936}.
%
%Multi-view \gls{CCA}, a generalisation of two-view \gls{CCA}, extends the above formulation to~\(m\) views, learning projections \( \proj_i \in \real^{n_i \times k}, i= 1, \dots, m \).
%
Projections are obtained via a generalised eigenvalue decomposition, \( A \bm{v} = \lambda B\bm{v}\)~\citep{kettenringMultiCCA1971,hardoonMultiCCA2004,bachkernelICA2002}, where \(A\) and \(B\) are the inter- and intra-view correlation matrices.
Projection matrix \(W_i \in \real^{n_i \times k}\)  embeds~\(\bfx_i\) from view~\(i\) as
\(\phi(\bfx_i; W_i) = D_{\bm{\lambda}}^p W_i^{\top} \bfx_i\),  {}% \in \mathbb{R}^k
where \(D_\lambda\) is a diagonal matrix of the top \(k\) (sorted) eigenvalues~\(\bm{\lambda}\), and~\(p\) is a scalar weight.
With \gls{CCA}, ranking and retrieval across views~\(\{ \bfx_i, \bfx_j \}\) is
performed by computing correlation between projections~\(\mathrm{corr}(\bfx_i, \bfx_j) = \frac{\psi(\bfx_i)^\top \psi(\bfx_j) }{\norm{\psi(\bfx_i)}_2 \norm{\psi(\bfx_j)}_2}\)
where \(\psi\) is a mean-centred (over \emph{train} set) version of \(\phi\).

% \vspace*{-0.5em}
% \paragraph{Scoring/generating answers with \gls{CCA}}
% \label{sec:prelim-cca2}

Using \gls{CCA}, learnt embeddings between answers and questions (\gls{CCA}\textsc{-aq}) are used to compute the ranking and \gls{NDCG} metrics.
%
%A schematic of this is shown in \cref{fig:schema}.
%
\gls{CCA} can also be used to \emph{generate} answers using correlations.
For a given test question, its \(100\) nearest-neighbour questions (based on correlation under the A-Q model) are extracted from the train set.
Their \(100\) corresponding ground-truth answers are used to construct a pseudo-candidate set.
Answers are \emph{generated} by the model, denoted \gls{CCA}\textsc{-aq-g}, by sampling from this set in proportion to correlation with the test question (see Figure 3 in~\citet{massiceti2018novd}).

\section{\!\!Shortfalls of Current \textit{VisDial} Evaluation\!\!}
\label{sec:shortcomings}

The fact that a simple, lightweight \gls{CCA} model performs favourably in \gls{MR} with current \gls{SOTA} models, while completely ignoring the image and dialogue history, and requiring an order of magnitude fewer learnable parameters and mere seconds on \textsc{cpu} to train, is a cause for concern.
Not only do prior results suggest that implicit correlations between just the questions and answers exist in the data (see~\cref{fig:another-vd-fail}), but also that the current evaluation scheme generally is not flexible enough to account for variation in answers to visually-grounded questions.
%
%rendering it possible to perform \gls{VD} without heed to either the visual or the sequential dialogue aspects.
%
Here we summarise the existing evaluation scheme, discuss the hidden factors affecting it, and make the case that to better capture a model's performance on the \gls{VD} task, there must be changes to the evaluation scheme.

\subsection{Current evaluation scheme}

Given a test question, the current \textit{VisDial} evaluation relies on ranking its candidate answers~\citep{Das_VisualDialog}, derived from scoring the answers under the trained (generative) model's likelihood (see~\cref{sec:prelim-vd}).

A suite of rank-based metrics is then computed:
\acrfull{MR} and \gls{MRR} of the ground truth (\textsc{gt}) answer over data, and the average recall, measuring how often the \textsc{gt} answer falls within the top 1, 5, and 10 ranks, respectively.
%
% \begin{inparaenum}[1)]
% \item the \acrfull{MR} over the dataset of the ground-truth (\textsc{gt}) answer
% \item the inverse harmonic mean of the \textsc{gt} answer ranks---\acrlong{MRR} (\acrshort{MRR}); and
% \item the average recall at 1, 5, and 10, measuring how often the \textsc{gt} answer falls within the top 1, 5, and 10 ranks, respectively.
% \end{inparaenum}
%
These {\bf single-candidate} (i.e. \textsc{gt}) ranking metrics have been the norm since \textit{VisDial}'s inception.

A subsequent extension of the dataset (\textit{v1.0}) tasked 4-5 human annotators with labelling whether each answer in a candidate set is valid for a given image-question (a hard 0/1 choice) for a subset of the train and validation sets, denoted \(\mathcal{H}_t\) and \(\mathcal{H}_v\), respectively.
For each candidate answer~\(A\), the mean judgement across annotators becomes a \emph{relevance} score~\(\rho(A) \in [0,1]\).
A modified {\bf multi-candidate} ranking metric, the \gls{NDCG}, is then introduced: candidate answers' ranks are weighted by their relevance scores, excluding irrelevant~(\(\rho(A) = 0\)) answers.
See \cref{sec:app-ndcg} for further details.

\subsection{Analysing current shortfalls}

The above evaluation metrics, by construction, are not flexible enough to account for the many ways a question can satisfactorily be answered.
This limitation manifests in both the single- and multi-candidate ranking metrics, and hampers the measurement of a model's true ability to answer a visual question.
The limitation stems from:
\begin{compactenum}
\item ranking candidate sets that are ill-constructed for the ranking task, and
\item disregarding answers \emph{generated} by a model in favour of indirectly ranking these fixed sets.
\end{compactenum}

\subsubsection{Ranking ill-constructed candidate sets}

Candidate answer sets in \textit{VisDial} are typically observed to contain multiple feasible answers---as they include up to 50 nearest-neighbour answers \citep{Das_VisualDialog} to~\(A_{\text{gt}}\) in \textit{GloVe}~\citep{pennington2014glove} space.
Rank-based metrics, which assume a meaningful ordering of answers, are less informative when considering feasible-answer subsets.

We explicitly verify this characteristic of candidate answers using correlation, through the following experiment which learns a \gls{CCA} model between the question and answer features.
Computing the correlation between~\(A_{\text{gt}}\) and~\(A \in \bm{A} \setminus A_{\text{gt}}\), giving~\(\mathcal{C} = (\phi(A_{\text{gt}}, A_1), \ldots, \phi(A_{\text{gt}}, A_{100}))\), we then select the cluster of answers with correlations in~\([\mathcal{C}_{\text{max}}-\sigma, \mathcal{C}_{\text{max}}], \text{where\,\,} \mathcal{C}_{\text{max}}\!=\!\mathrm{max}(\mathcal{C}), \sigma\!=\!\mathrm{stdev}(\mathcal{C})\), roughly estimating answers which are plausibly similar to~\(A_{\text{gt}}\).
Given this cluster, we compute the mean and standard deviation of the correlations, as well as the cluster size, to measure how small and tightly packed these clusters are.
We average these across all candidate sets, giving an average mean correlation of \(0.58\), an average standard deviation of \(0.22\), and an average cluster size of \(12.30\).
These results support the idea that an \emph{equivalence class} of feasible answers exist within each candidate set, which can then adversely affect both classes of metrics described below.

{\bf Single-candidate ranking metrics} assign a single answer, the labelled \textsc{gt}, as the \emph{only} correct answer in the candidate set, and are purely a function of this privileged answer's rank.
As a result, these metrics unduly penalise models that rank alternate, but equally feasible, answers highly.
\Gls{MR}, \gls{MRR}, and \textsc{r@1,5,10} are thus only weakly indicative of performance on the \gls{VD} task, and are unable to differentiate between equally good models.

\begin{figure}
\begin{subfigure}{.22\textwidth}
    \includegraphics[width=\textwidth]{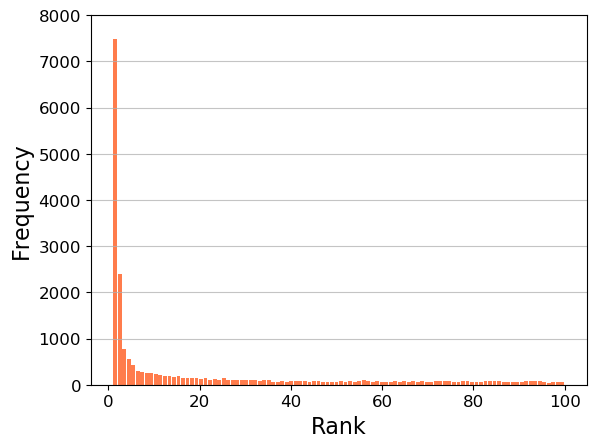}
    \vspace*{-2em}
    \caption{\textsc{hrea-qih-g}}
  \end{subfigure}
  %%%%%%%%%%%%%%
  \begin{subfigure}{.22\textwidth}
    \includegraphics[width=\textwidth]{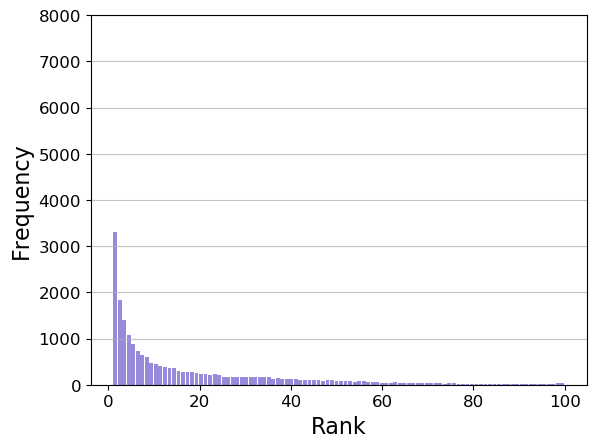}
    \vspace*{-2em}
    \caption{\gls{CCA}\textsc{-aq}}
  \end{subfigure}
\vspace*{1em}
\caption{%
  Distribution of \textsc{gt} answer ranks across \textit{VisDial v1.0} val set. Highly-parametrised \gls{SOTA} models (a) pushes the \textsc{gt} rank towards 1, ignoring other potentially feasible answers, in contrast to \gls{CCA}\textsc{-aq} (b).
  %
  % This adversely affects \gls{MRR}/\textsc{Recall@1,5,10}, and causes non-diverse answer generations.
}
\label{fig:histo}
\vspace*{-0.1em}
\end{figure}

The ill-constructed candidate sets also render single-candidate metrics unable to \emph{rule out} poor models.
In other words, models with poor \gls{MR}, \gls{MRR} and \textsc{r@1,5,10} aren't necessarily poor at \gls{VD}.
This is markedly the case for \gls{MRR} and \textsc{r@1,5,10} which are, by definition\footnote{While obvious for recall, \gls{MRR} as the inverse harmonic mean, weighs smaller ranks more strongly than larger ranks.}, biased toward low ranks---a model predicting five \textsc{gt} answers at rank 1, and five at rank 10, scores better \gls{MRR}/\textsc{r@1,5,10} than a model with all ten \textsc{gt} answers at rank 2 (coincidentally, these results are meaningless if the candidate set contained \(10\) equally feasible answers).
This bias particularly affects models trained with a single-answer objective (i.e. all \gls{SOTA}) models.
To see why, we show the distribution of \textsc{gt} answer ranks between \gls{CCA}\textsc{-aq} and a \gls{SOTA} model in~\cref{fig:histo}.
The \gls{SOTA} model is skewed toward the \textsc{gt} answer achieving rank 1---the combined result of a single-answer objective and high parametrisation.
This leads \gls{SOTA} models to view other feasible answers in the set as no different if ranked 2 or 100.
\gls{CCA}\textsc{-aq} by contrast ignores rank and simply learns by maximising A-Q correlation, likely leading it to rank other feasible answers highly.
Thus, models favouring low ranks by virtue of their learning objective may achieve better \gls{MRR}/\textsc{r@1,5,10}, but not be discernibly better than models accounting for multiple answers being correct.
% this explains why cca does well on MR but not other metrics in~\citep{massiceti2018novd}
%

These findings, together, suggest that the single-candidate metrics cannot reliably quantify performance and compare models in lieu of the \gls{VD} task.

{\bf Multi-candidate ranking metrics}, or \gls{NDCG}, undoubtedly take a step in the right direction by forgoing just a single correct answer, and weighting the predicted ranking with human-annotated relevance scores for multiple answers.
\gls{NDCG}, however, is still a function of a ranking, and hence assumes that a single optimal ordering of candidate answers exists.
The presence of multiple \emph{equally} feasible answers in the candidate sets thus breaks this assumption and can skew the \gls{NDCG}, albeit to a lesser degree than \gls{MR}, \gls{MRR}, and \textsc{r@1,5,10}.
%
%%% give example

Moreover, the degree of answer similarity within these subsets raises further concerns for the reliable computation of \gls{NDCG}.
Requiring annotation of 100 valid (i.e. similar) answers is an arduous task, and converting hard 0/1 judgements into relevance scores over just a handful (4-5) of annotators can be noisy.
Our analysis reveals the following quirks:
\begin{compactitem}
\item \(18.15\%\)\big/\(47.14\%\) of the validation/train annotated subsets (\(\mathcal{H}_v\)\big/ \(\mathcal{H}_t\)), do not have a {\em single} candidate answer with relevance score \(1.0\), not even the ground-truth, indicating poor consensus.
%363/2000, 973/2064
\item 20.69\%\big/9.01\% of samples, respectively, consider the ground-truth irrelevant (\(\rho(\gtA) = 0\)).
\end{compactitem}
% This is evident upon qualitative inspection (see \cref{fig:bad-ndcg})
% \missingfigure[figwidth=\columnwidth]{\label{fig:bad-ndcg}figure showing failure cases of NDCG - spurious annotations, and variance of NDCG scores}

Coupled with this, the scale of \textit{VisDial} makes obtaining annotations a daunting (and expensive) task---reflected in the fact that only a small fraction of the data, one question per image, has annotations (see \cref{sec:prelim}), which implies evaluations effectively ignore dialogue history.
% %
% This means that the current evaluation ignores the notion of dialogue in the visual \emph{dialogue} task.
%
Also, without more annotators (and hence cost/time), obtaining relevance scores at-scale may well be meaningless.
%even if the round of dialogue for which one has judgments is uniformly distributed.
%

\subsubsection{Ignoring generated answers}

The ultimate goal of \gls{VD} is to produce an answer to a given question, not to pick an answer from a set---our primary motivation for focussing on the \emph{generative} \gls{VD} task.
The current evaluation, rather than directly evaluating the answers generated by a model, evaluates by how well a model ranks a fixed set of candidate answers.
Not only is this problematic because of the candidate sets' limitations (as described above), but also because it:
\begin{inparaenum}[1)]
\item disregards diversity in answer generations, a necessary feature for a human-like answering agent, and
\item goes against established practice in the \acrshort{VQA} literature~\citep{Antol_VQA} which evaluates by comparing the predicted answer to answers collected from 10 human annotators.
\end{inparaenum}
While it is expected that scoring a valid answer by its likelihood is a reasonable measure of a model's ability to \emph{generate} a good answer, this may not necessarily be the case when there are multiple potential answers, some not even in the candidate set.
Although likelihoods can serve as a \emph{relative} measure between candidates, the highest-probability answers may be entirely different or unrelated---indicating a poorly learnt model.
This supports the idea that a metric which ignores generated answers may fail to account for models no less ``good'' at the \gls{VD} task.

\section{A Revised Evaluation for \textit{VisDial}}
\label{sec:new-eval}

The analysis in \cref{sec:shortcomings} indicates that an evaluation well matched to the underlying goals of \gls{VD} should:
\begin{compactenum}[i)]
\item directly use answers generated by the model, \label{feat:gen}
\item account for multiple valid answers, and \label{feat:multi}
\item do the above at scale over the entire dataset. \label{feat:scale}
\end{compactenum}
We thus develop a revised evaluation scheme for \gls{VD} which meets these three criteria.
Its basis lies in measuring how similar an answer \textit{generated} by a given model is to a \textit{set} of feasible reference answers for a given question and image.
We describe similarity quantification in \cref{sec:measuring-similarity} and the construction of high-quality reference sets in \cref{sec:obtaining-refs}.

\subsection{Measuring similarity} \label{sec:measuring-similarity}
We measure similarity using established \textsc{nlp} consensus metrics  between a predicted answer and a reference set of valid answers.
Crucially, the predicted answer is \textit{generated} by the model directly, and the reference set contains more than one element, accounting for the presence of multiple valid answers.
We use two classes of metric for capturing consensus: overlap and embedding distance.

{\bf Overlap-based} metrics compute the overlap or co-occurrence of \(n\)-grams (word couplets of size~\(n\)) between pairs of sentences---here, the generated answer and each answer in the reference set.
%
% We then compute the average over the reference set.
%
We use two such metrics: {\sc cider} \citep{Vedantam_2015cider} and {\sc meteor} \citep{Denkowski_2014meteor},
%(designed to address the limitations of {\sc bleu}~\citep{Papineni_2002bleu})
motivated by their extensive use in image captioning benchmarks~\citep{ChenFLVGDZ15,hodosh2013framing,young2014image}.
Both are known to be well correlated with human judgements.
{\sc cider} computes the cosine similarity between a pair of vectors, each of which is composed of the term-frequency inverse-document-frequencies (tf-idf) of the sentence's \(n\)-grams.
For \(0 < n \leq i\), similarities are averaged over all \(n\)-grams up to length \(i\).
{\sc meteor} is similar, but first applies a uni-gram matching function, before computing a weighted harmonic mean between uni-gram precision and recall, with a fragmentation penalty on the matching.

{\bf Embedding distance-based} metrics arise from a rich literature in capturing semantic similarity between natural language expressions~\citep{bojanowski2017enriching,pennington2014glove,devlin2018bert,peters2018elmo,sharma2017relevance}.
Motivated by the recent successes of {\sc bert}~\citep{devlin2018bert} and {\it FastText}~\citep{bojanowski2017enriching} in a variety of \textsc{nlp} tasks, we use each method to embed the generated answer and each reference set answer, computing the \(L_2\) and cosine similarity (CS) between them, averaging over the reference set.
The embedding metrics aim to complement the overlap-based metrics and guard against limitations of the latter that might arise due to answer lengths (one- or two-word) frequently seen in the \acrshort{VD} data.

\subsection{Obtaining answer reference sets} \label{sec:obtaining-refs}
We now describe how to obtain reference sets for the similarity measures defined above.

\subsubsection{Using humans} \label{sec:obtaining-refs-humans}
For a small subset of the \textit{VisDial} validation set, \(\mathcal{H}_v\), soft relevance scores are available (from human annotators) for each of the 100 candidate answers associated with each image-question (see~\cref{sec:prelim-vd}).
Using these scores, we construct answer reference sets for each image-question, composed of all the candidate set answers deemed valid by at least one annotator, i.e.\ \(\rho(A) > 0\), where to our surprise, we found multiple instances where \(\rho(\gtA) = 0\).
Protecting against such cases, we define the human-annotated reference set \(H = \{A : \rho(A) > 0, \forall A \in \bm{A}\} \cup \{\gtA\}\).

\subsubsection{Using semi-supervision at scale} \label{sec:obtaining-refs-at-scale}
Human-annotated relevance scores, and hence reference sets, however, are available for only a fraction of the dataset---less than 1\% of questions!
The scale of \textit{VisDial}---on the order of \(10^6\) questions, each with \(100\) candidate answers---makes extending these annotations to the entire dataset extremely challenging.
Assuming \(\$0.05\) per question, each presented to \(10\) workers, would incur a cost of over \(\$500,000\) and substantial annotation time!

We therefore propose a semi-supervised approach which harnesses the annotations we \emph{do} have: given a candidate set of answers for an image-question, we learn to extract the valid answers, and hence automatically construct a reference set.
Not only does this enable us to obtain reference sets at scale, but it also circumvents the time, cost and idiosyncrasies associated with human annotation.

Our approach is based on \gls{CCA}, and uses the relevance-annotated subset of the full train set, \(\mathcal{H}_t\), as training data.
Similar to~\cref{sec:obtaining-refs-humans}, we construct \emph{training} reference sets \(H_t\) using \(\mathcal{H}_t\).
Pairing each question with \emph{all} answers in \(H_t\), we learn a \gls{CCA} model between the questions and answers.
With this model, denoted \gls{CCA}\textsc{-aq}\textsuperscript{*}, we compute correlations between~\(\gtA\) and \(\bm{A} \setminus \gtA\), giving \(\mathcal{C} = (\phi(\gtA, A_1), \ldots, \phi(\gtA,A_{100}))\) similar to \cref{sec:shortcomings}.
We then cluster these correlations in \(\mathcal{C}\) to construct a reference set~\(\Sigma = \{A : \phi(\gtA, A) \in [\mathcal{C}_{\text{max}} - \sigma, \mathcal{C}_{\text{max}}]\} \cup \{\gtA\} \)
where \(\mathcal{C}_{\text{max}} = \mathrm{max}(\mathcal{C})\), and \(\sigma = \mathrm{stdev}(\mathcal{C})\).
% We then cluster the correlations in \(\mathcal{C}\) to construct a reference set~\(\Sigma\) as:
% %
% \begin{align} \nonumber
% \Sigma = &\{A : \phi(\gtA, A) \in [\mathcal{C}_{\text{max}} - \sigma, \mathcal{C}_{\text{max}}]\} \cup \{\gtA\} \\
% & \text{where } \mathcal{C}_{\text{max}} = \mathrm{max}(\mathcal{C}), \quad \sigma = \mathrm{stdev}(\mathcal{C}) \nonumber
% \end{align}
%
Intuitively, this extracts the cluster of answers with highest correlation or similarity to the ground-truth answer.
With this semi-supervised approach, we easily and quickly obtain reference sets at scale for the \textit{entire} \textit{VisDial} dataset.

\paragraph{Verifying automatic reference sets}
The validity of the revised evaluation is contingent on the validity of the automatic reference sets---that they are composed of valid answers.
We verify this by:
\begin{compactenum}[(1)]
\item computing intersection metrics between the human-annotated and automatic reference sets, %in \(\mathcal{H}_v\),
\item using \gls{AMT} to verify the sets, and
\item measuring how \textit{training} a \gls{VD} model on these sets can improve performance on \gls{VD}.
\end{compactenum}

For (1), we compute the \gls{IOU}, precision, recall, and set size of the automatic~\(\Sigma\) and human-annotated~\(H\) reference sets on the validation subset~\(\mathcal{H}_v\) (\cref{tab:clusters}).
These metrics serve as a simple heuristic and we use them to compare clustering methods (see extended comparison in~\cref{sec:app-clustering}).
Our best method, \(\Sigma\), extracts similar sized clusters to \(H\) (\(7.17\) vs \(12.77\)) with good precision (\(62.48\%\); i.e. it selects answers maximally in~\(H\)), supporting the similarity of \(\Sigma\) and \(H\).
\begin{table}[t]
  \centering
  \def\s{\hspace*{7pt}}
  \scalebox{0.7}{
  \begin{tabular}{c@{\s}c@{\s}c@{\s}c@{\s}c@{\s}}
    \toprule
    \(C\) & \(\frac{|H \cap C|}{|H \cup C|}\) & \(\frac{|H \cap C|}{|C|}\) & \(\frac{|H \cap C|}{|H|}\) & \(|C|\) \\
    \midrule
    \(H\)       & 100.00 (0.00) & 100.00 (0.00) & 100.00 (0.00) & 12.77 (7.24) \\[0.5ex]
    \(\Sigma\)  & 24.13 (16.73) & 62.48 (31.24) & 32.91 (23.52) & 7.17 (6.94) \\
    %\(M\)       & 25.01 (18.40) & 59.19 (31.55) & 39.35 (28.86) & 12.00 (16.07) \\
    %\(G\) (n=5) & 25.59 (17.69) & 59.20 (30.71) & 35.74 (24.47) & 7.91 (6.18) \\
    \bottomrule
    \end{tabular}}
  \vspace*{-1ex}
  \caption{%
    Evaluation of intersection metrics computed on human-annotated reference sets~\(H\) and automatic reference sets~\(\Sigma\), on the validation subset~\(\mathcal{H}_v\).
    Values in parentheses denote standard deviation across the set.}
  \label{tab:clusters}
  \vspace*{1ex}
\end{table}

For (2), we turn to \gls{AMT}.
Given an image, question and answer reference set (from either \(\Sigma\) or \(H\)) as a task, a turker is asked to de-select all infeasible answers (see~\cref{sec:app-clustering} for \gls{AMT} user interface).
For each task, scores are averaged over 5 turkers.
We then measure the proportion of the reference set selected, and the proportion of the set where at least 1 turker selected each answer.
For a subset of tasks randomly sampled from \(\mathcal{H}_v\) or the full validation set, in~\cref{tab:mturk-cluster-val} we observe that our proposed semi-supervised reference sets are similar to the ones obtained using humans.
\begin{table}[t]
  \centering
  \def\s{\hspace*{7pt}}
  \scalebox{0.8}{
  \begin{tabular}{r@{\s}c@{\s}c@{\s}}
    \toprule
    & \(C=H\) & \(C=\Sigma\) \\
    \cmidrule{2-3}
    \# tasks & \(1,680\) & \(5,040\) \\
    \# turkers per task & 5 & 5 \\
    \% \(C\) selected & \(81.48 \;(2.15)\) & \( 70.66 \;(5.45) \) \\
    \% \(C\) selected (\(\ge 1\) turker) & \(98.80 \;(0.37)\) & \( 95.52 \;(2.22) \) \\
    \bottomrule
    \end{tabular}}
  \vspace*{-1ex}
  \caption{%
    \Acrshort{AMT} validation of automatic reference sets \(\Sigma\) against human-annotated sets \(H\).
    For each task, given an image, question and answer set (from either \(\Sigma\) or \(H\)), turkers are asked to deselect infeasible answers, with scores averaged over 5 turkers.
    We measure the proportion of each set selected, and the proportion of each set where \(\geq\)1 answer  was selected. Variance in brackets. }
  \label{tab:mturk-cluster-val}
  \vspace*{1ex}
\end{table}

Finally, in (3), we intuit that if reference sets \(\Sigma\) contain answers similar to the correct answer, then a model trained on only these sets should improve performance on the \gls{VD} task.
We, therefore, pair each question in the training subset~\(\mathcal{H}_t\) with each of the answers in its corresponding~\(\Sigma\), and train a \gls{CCA}\textsc{-aq} model.
As a baseline, we repeat this experiment, but pairing the questions with answers from~\(H\) instead of~\(\Sigma\).
We show in~\cref{tab:ndcg-cluster} (top 2 rows), the model trained using~\(\Sigma\) performs better than that employing the human-annotated reference sets~\(H\) across the battery of ranking metrics, including \gls{NDCG}.
As a further check, we train a \gls{CCA}\textsc{-aq} model on~\(\mathcal{H}_t\), but only between questions and their single ground-truth answers~\(\gtA\) (as opposed to {\em all} answers in \(H\) or \(\Sigma\)).
As we address in \cref{sec:shortcomings}, this model surprisingly outperforms the baseline using~\(H\) as reference on the single-candidate ranking metrics, however, as expected, \gls{NDCG} paints a better picture, showing reduced performance.
Finally we conduct (3) across the whole dataset, learning a \gls{CCA}\textsc{-aq} model using~\(\Sigma\), over the entire training data of \textit{VisDial (v1.0)}.
The last two rows of \cref{tab:ndcg-cluster} compare this model against the standard \gls{CCA}\textsc{-aq} trained on questions and ground-truth answers.
We observe a substantial improvement in \gls{NDCG}, with what is effectively a simple data augmentation procedure using~\(\Sigma\).
This three-part verification supports the existence of valid answers in the automated reference set, which subsequently supports our revised evaluation scheme.
\begin{table}[t]
  \centering
  \def\s{\hspace*{4ex}}
  \newcommand{\mr}[2]{\multirow{#1}{*}{{#2}}}
  \newcommand{\mc}[2]{\multicolumn{#1}{c}{{#2}}}
  \scalebox{0.65}{
  \setlength{\tabcolsep}{3pt}
  \begin{tabular}{@{}cr>{$}c<{$}@{\s}*{6}{c}@{}}
    \toprule
    \mc{2}{Train}               & \mr{2}{Ref} & \mr{2}{MR} & \mr{2}{R@1} & \mr{2}{R@5} & \mr{2}{R@10} & \mr{2}{MRR} & \mr{2}{NDCG} \\
    \cmidrule{1-2}
    \text{Set}     & \#QA pairs &             & \(\downarrow\) & \(\uparrow\) & \(\uparrow\) & \(\uparrow\) & \(\uparrow\) & \(\uparrow\) \\
    \midrule
    \mr{3}{\(\mathcal{H}_t\)}
      & 15,317     & H         & 26.49      & 6.05        & 21.50       & 35.53        & 0.1550      & 0.3647 \\
      & 17,055     & \Sigma    & {\bf 20.36}      & 8.35        & 32.88       & {\bf 48.78}        & 0.2066      & {\bf 0.3715} \\
      & 1996       & \{\gtA\}  & 23.71      & {\bf 13.13}       & {\bf 34.05}       & 46.90        & {\bf 0.2428}      & 0.2734 \\
    \cmidrule(l){2-9}
    \mr{2}{all}
      & 10,419,489 & \Sigma    & 17.20      & 10.73       & 34.20       & 51.80        & 0.2312      & {\bf 0.4023} \\
      & 1,232,870  & \{\gtA\}  & {\bf 17.07}      & {\bf 16.18}       & {\bf 40.18}       & {\bf 55.35}        & {\bf 0.2845}      & 0.3493 \\
    \bottomrule
  \end{tabular}}
  \vspace*{-1ex}
  \caption
  {%
    Evaluating the utility of automated reference sets~\(\Sigma\) on standard \gls{VD} evaluation.
    \gls{CCA}\textsc{-aq} models were trained on the indicated subsets (\(\mathcal{H}_t\) or all) of \textit{VisDial (v1.0)}, with answers from different sets (`Ref'), and tested on the \href{https://visualdialog.org/challenge/2018}{evaluation test server} to compute standard metrics. Arrows indicate which direction is better.
  }
  \label{tab:ndcg-cluster}
\end{table}

\section{Experimental Analyses}

Here we include experimental analyses, focussing in particular on the performance of models under our revised evaluation schemes discussed in~\cref{sec:new-eval}.

% \subsection{Experimental details}
%
% \paragraph{\gls{CCA} feature extractors}%
%
%For the images, we employ pre-trained \textit{ResNet34}~\citep{He_2016deep}, extracting a 512-dimensional feature---the output of the \emph{avg pool} layer after \emph{conv5}.
%
We represent words in the questions/answers as 300-dimensional \textit{FastText}~\citep{bojanowski2017enriching} embeddings.
To obtain sentence embeddings, we simply average word embeddings following generally received intuition~\citep{aurora2017simple,wieting2017random}, padding or truncating up to 16 words following~\citet{massiceti2018novd}.
%
% \paragraph{\gls{SOTA} models}
%
We generate answers from \gls{CCA}\textsc{-aq-g} and the following \gls{SOTA} models: \textsc{hrea-qih-g}~\citep{Das_VisualDialog}, \textsc{hciae-dis-g}~\citep{lu2017best} and \textsc{rva}~\citep{niu2019recursive}, with * indicates use of beam search.
For each, we train on the full \textit{VisDial v1.0} train set, cross-validate on \gls{MRR}, and select the best epoch's model for subsequent evaluation.

\subsection{Revised evaluation results}

\paragraph{Testing on \(\mathcal{H}_v\)}
\Cref{tab:consensus} (left) shows the overlap and embedding distance scores of answers generated by models, measured against human-annotated reference sets \(H\) for the validation subset \(\mathcal{H}_v\).
Note, we report on~\(\mathcal{H}_v\) because relevance scores are available for only part of \textit{VisDial}'s full validation set and are publicly unavailable for its test set.

\begin{table*}[!th]
\def\s{\hspace*{3ex}}
\def\mc#1#2{\multicolumn{#1}{c}{{\sc #2}}}
\def\up{\(\uparrow\)}
\def\down{\(\downarrow\)}
\def\k1{\(^{\star}\)\!\!}
\begin{subtable}{0.52\textwidth}
\centering
\scalebox{0.53}{
\setlength{\tabcolsep}{3pt}
\begin{tabular}{@{}l*{4}{c}@{\s}c@{\s}*{2}{c}@{\s}*{2}{c}@{}}
  \toprule
  Model                  & \mc{4}{cider\up}                  & {\sc meteor\up} & \mc{2}{bert}    & \mc{2}{FastText} \\
                         \cmidrule(r){2-5}                                        \cmidrule(l){7-8}\cmidrule(l){9-10}
                         & n=1    & n=2    & n=3    & n=4    &                 & \(L_2\)\down & CS\up          & \(L_2\)\down & CS\up      \\
  \midrule
  \(\Gamma_H\)             & 0.2765 & 0.2151 & 0.1810 & 0.1513 & 1.0000          &  4.7000 & 0.9334      & 1.8757 & 0.6992  \\
  \cmidrule{2-10}
  \gls{CCA}\textsc{-aq-g}  & 0.0721 & 0.0434 & 0.0298 & 0.0226 & 0.2713 & 7.1231 &  0.8690 &  3.1251 &  0.4555 \\
                           %& (0.1187) & (0.0726) & (0.0506) & (0.0389) & (0.3083) & (1.7118) &  (0.0626) &  (0.8266) &  (0.1364) \\
  \textsc{hrea-qih-g}     & 0.0880 & 0.0483 & 0.0333 & 0.0252 & 0.4813 & 6.2875 &  0.8927 &  2.9724 &  0.5079  \\
  %\hfill \(\sigma\)               & (0.1161) & (0.0671) & (0.0480) & (0.0374) & (0.4264) & (1.5281) &  (0.0536) &  (0.7308) &  (0.1146)  \\
  \textsc{hrea-qih-g*}    & {\bf 0.1359} & {\bf 0.0721} & {\bf 0.0494} & {\bf 0.0372} & {\bf 0.7149} & {\bf 5.5727} & {\bf 0.9149}  &  3.2664 &  0.4971 \\
  %\hfill \(\sigma\)               & (0.1431) & (0.0789) & (0.0564) & (0.0430) & (0.4291) & (1.0037) &  (0.0278)  &  (0.6524) &  (0.1148) \\
  %\textsc{mna-qih-g*}    & {\bf 0.1365} & 0.0713 & 0.0484 & 0.0364 & {\bf 0.7316} & {\bf 5.5549} &  {\bf 0.9153} &  3.3271 &  0.4920 \\
  %\hfill \(\sigma\)      &(0.1394) & (0.0767) & (0.0540) & (0.0405) & (0.4274) & (1.0043) &  (0.0279) &  (0.6399) &  (0.1153) \\
  \textsc{hciae-g-dis} & 0.1338 & 0.0718 & 0.0493 & {\bf 0.0372} & 0.6758 & 5.6690 &  0.9122 &  3.1551 &  0.5049 \\
  %\hfill \(\sigma\) & (0.1478) & (0.0809) & (0.0564) & (0.0426) & (0.4291) & (1.0671) &  (0.0304) &  (0.6940) &  (0.1158) \\
  \textsc{rva}            & 0.1042 & 0.0563 & 0.0385 & 0.0291 & 0.5328 & 6.1466 &  0.8967 &  {
\bf 2.9543} &  {\bf 0.5161} \\
  %\hfill \(\sigma\)     &(0.1311) & (0.0709) & (0.0491) & (0.0371) & (0.4304) & (1.4871) &  (0.0509) &  (0.7253) &  (0.1170) \\
  \bottomrule
\end{tabular}}
\label{tab:consensus-human}
\end{subtable}%
\hspace*{0.5em}
\begin{subtable}{0.35\textwidth}
\centering
\scalebox{0.53}{
\setlength{\tabcolsep}{3pt}
\begin{tabular}{@{}l*{4}{c}@{\s}c@{\s}*{2}{c}@{\s}*{2}{c}@{}}
  \toprule
  & \mc{4}{cider\up}                  & {\sc meteor\up} & \mc{2}{bert}    & \mc{2}{FastText} \\
                         \cmidrule(lr){2-5}                                                   \cmidrule(lr){7-8}\cmidrule(lr){9-10}
                        & n=1    & n=2    & n=3    & n=4    &              & \(L_2\)\down& CS\up     & \(L_2\)\down& CS\up      \\
  \midrule
  % \(\gtA\)               & 0.3502 & 0.2479 & 0.2004 & 0.1692 & 0.9955       & \multicolumn{4}{c}{\rule[2pt]{25ex}{0.5pt}} \\ % 4.8643 & 0.9215 & 2.2155 &  0.6764 \\
  \(\Gamma_\Sigma\) & 0.4212 & 0.3429 & 0.2991 & 0.2583 & 1.0000 & 4.2891 & 0.9373 & 1.6518 & 0.7614 \\
  \cmidrule{2-10}
  & 0.0789 & 0.0461 & 0.0313 & 0.0235 & 0.1864 & 7.1873 &  0.8673 &  3.0908 &  0.4782  \\
  %\hfill \(\sigma\)               & (0.1634) & (0.1040) & (0.0716) & (0.0544) & (0.2478) & (1.6917) &  (0.0610) &  (0.9044) &  (0.1639)\\
  & 0.1109 & 0.0597 & 0.0409 & 0.0308 & 0.3710 & 6.2743 &  0.8924 &  {\bf 2.8815} &  0.5334 \\
  % \hfill \(\sigma\)               & (0.1591) & (0.0909) & (0.0647) & (0.0494) & (0.4098) & (1.6571) &  (0.0540) &  (0.7971) &  (0.1415) \\
  & 0.1580 & 0.0835 & 0.0568 & 0.0428 & {\bf 0.5269} & {\bf 5.7023} &  {\bf 0.9097} &  3.1888 &  0.5196 \\
  %\hfill \(\sigma\)                &(0.1824) & (0.1048) & (0.0752) & (0.0580) & (0.4654) & (1.4204) &  (0.0399) &  (0.7521) &  (0.1525) \\
  %& 0.1515 & 0.0777 & 0.0523 & 0.0393 & 0.5239 & 5.7056 &  {\bf 0.9099} &  3.2581 &  0.5119 \\
  %\hfill \(\sigma\) & (0.1756) & (0.0938) & (0.0649) & (0.0495) & (0.4697) & (1.4043) &  (0.0396) &  (0.7169) &  (0.1509) \\
  & {\bf 0.1614} & {\bf 0.0878} & {\bf 0.0605} & {\bf 0.0457} & 0.5138 & 5.7374 &  0.9087 &  3.0389 &  0.5347 \\
  %\hfill \(\sigma\) & (0.1859) & (0.1125) & (0.0827) & (0.0644) & (0.4497) & (0.2314) & (0.1716) & (0.0812) & (0.4533) & (1.4411) &  (0.0409) &  (0.7878) &  (0.1508) \\
  & 0.1209 & 0.0650 & 0.0445 & 0.0336 & 0.4033 & 6.1629 &  0.8956 &  2.9040 &  {\bf 0.5353} \\
  %\hfill \(\sigma\)              &(0.1644) & (0.0941) & (0.0678) & (0.0521) & (0.4238) & (1.6415) &  (0.0527) &  (0.7994) &  (0.1437) \\
  \bottomrule
\end{tabular}}
%\vspace*{-1ex}
\label{tab:consensus-auto}
\end{subtable}
\vspace*{-1.5ex}
\caption
{%
  Overlap and embedding distance metrics computed for \(k=1\) generation against human-annotated reference sets \(H\) on the validation subset \(\mathcal{H}_v\) (left), and automatic reference sets \(\Sigma\) on the \emph{entire} validation set (right).
  For \textsc{hrea-qih-g}, on average \(\sim\)6 answers are the empty string, which are excluded from the computation.
  Metrics marked~\(\uparrow\) indicate higher values are better, and those marked~\(\downarrow\) indicate lower values are better.
}
\label{tab:consensus}
\vspace*{1ex}
\end{table*}

We define a reference baseline for the overlap metrics, estimating upper bounds for the respective scores as \(\Gamma_H\), which cycles through answers in~\(H\), measures against~\(H\) itself and takes the maximum over the resulting scores.

\vspace*{-1ex}
\paragraph{Testing on whole dataset}

The final step is to use the validated automatic reference sets (from~\cref{sec:obtaining-refs}) to evaluate the models under the revised scheme for the complete \textit{VisDial (v1.0)} dataset.
\Cref{tab:consensus} (right) shows the overlap and embedding distance scores of answers generated by models, measured against the automatic reference sets~\(\Sigma\) for the whole validation set.
Again, we test on the validation set since ground-truth answers are not publicly available for the test set---something we require to construct~\(\Sigma\).
Note, the baseline~\(\Gamma_{\Sigma}\) here is different from before since the reference set is different: \(\Sigma\) instead of~\(H\).

\vspace*{-1ex}
\paragraph{Model comparison}
Comparing models which do \emph{not} employ beam search (i.e.\ no asterix), \textsc{hciae-dis-g} performs the best across all metrics except \textsc{FastText}, which \textsc{rva} wins (\cref{tab:consensus}).
This is consistent on \(\mathcal{H}_v\) \emph{and} the full validation set, despite \(\mathcal{H}_v\) being \(10\)-fold smaller---a further confirmation of \(\Sigma\)'s utility.
Note, results across all metrics are well below the reference baselines~\(\Gamma_H\big/\Gamma_\Sigma\), indicating there is still room for improvement.
Applying a beam-search on top of these models has the ability to further enrich the generations and improve performance on all metrics, as shown by \textsc{hrea-qih-g*}.
It is expected that applying a beam search to the best-performer \textsc{hciae-dis-g} would yield similar improvements.
Surprisingly, these results differ from the conclusions drawn from the rank-based evaluation (see~\cref{tab:full-cca-analysis} in supplement), where \textsc{rva} supersedes all other \gls{SOTA} models on \emph{all} rank metrics.
This suggests that just because a model can rank a single-ground truth answer highest, does not necessarily make it the best \emph{generative} \gls{VD} agent.
Our suite of overlap metrics and embedding distance metrics may help to explain \emph{why}.
For example, \textsc{cider} n=1 is a proxy for how well a model performs on one-word answers, which are highly prevalent in the dataset (e.g.\ ``Yes''/``No'').
\textsc{bert}, on the other hand, may help to measure generations with the closest semantic similarity to the reference sets.
Indeed, this is the sort of flexibility of purpose that is required when evaluating complex multi-modal tasks like \gls{VD}.

%
%Separation of evaluation across different \(n\)-gram lengths allows us to tease apart such characteristics, potentially providing further insight into models.

\begin{figure}[h]
\centering
\includegraphics[width=0.45\textwidth]{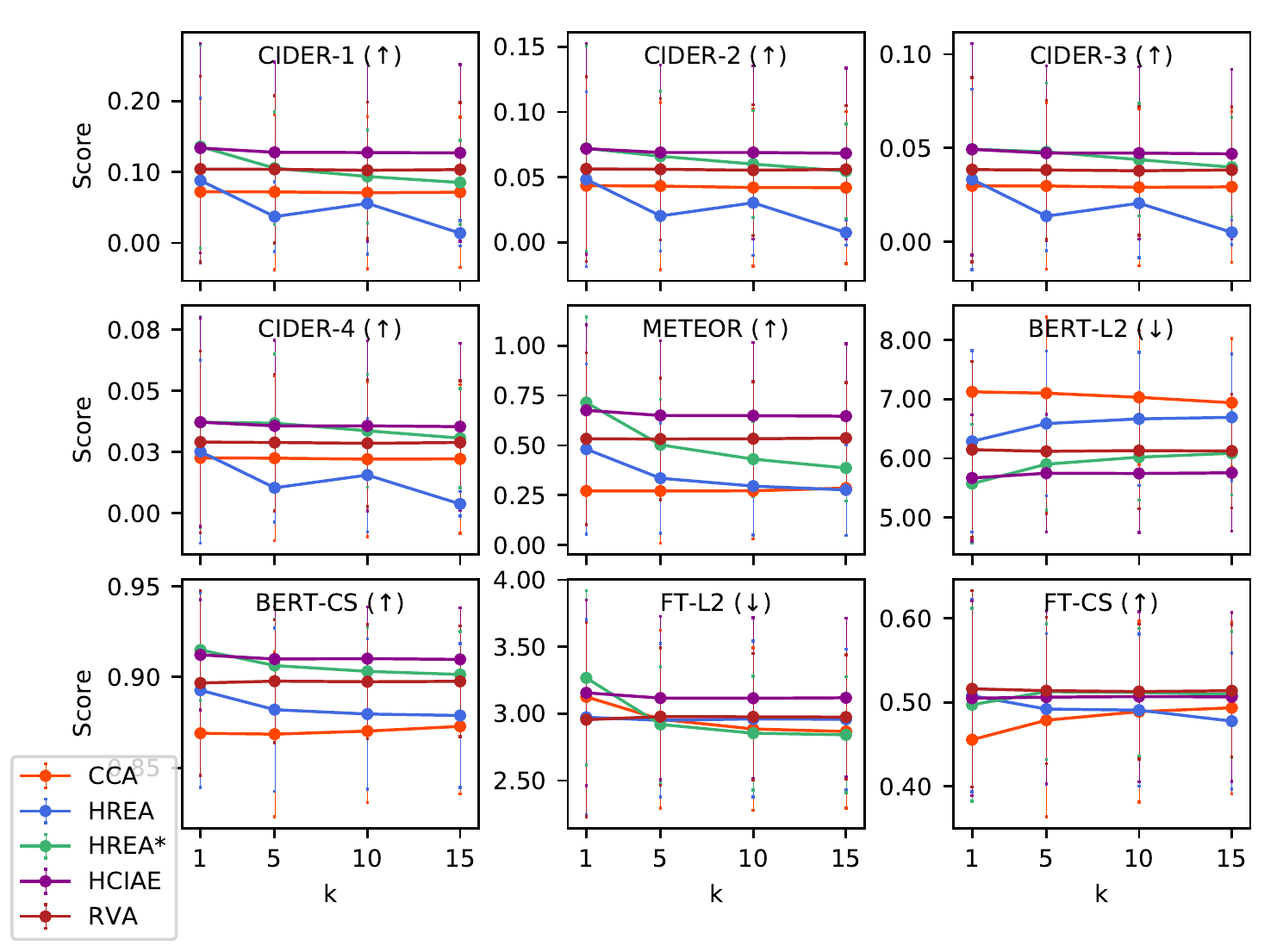}\\
\includegraphics[width=0.45\textwidth]{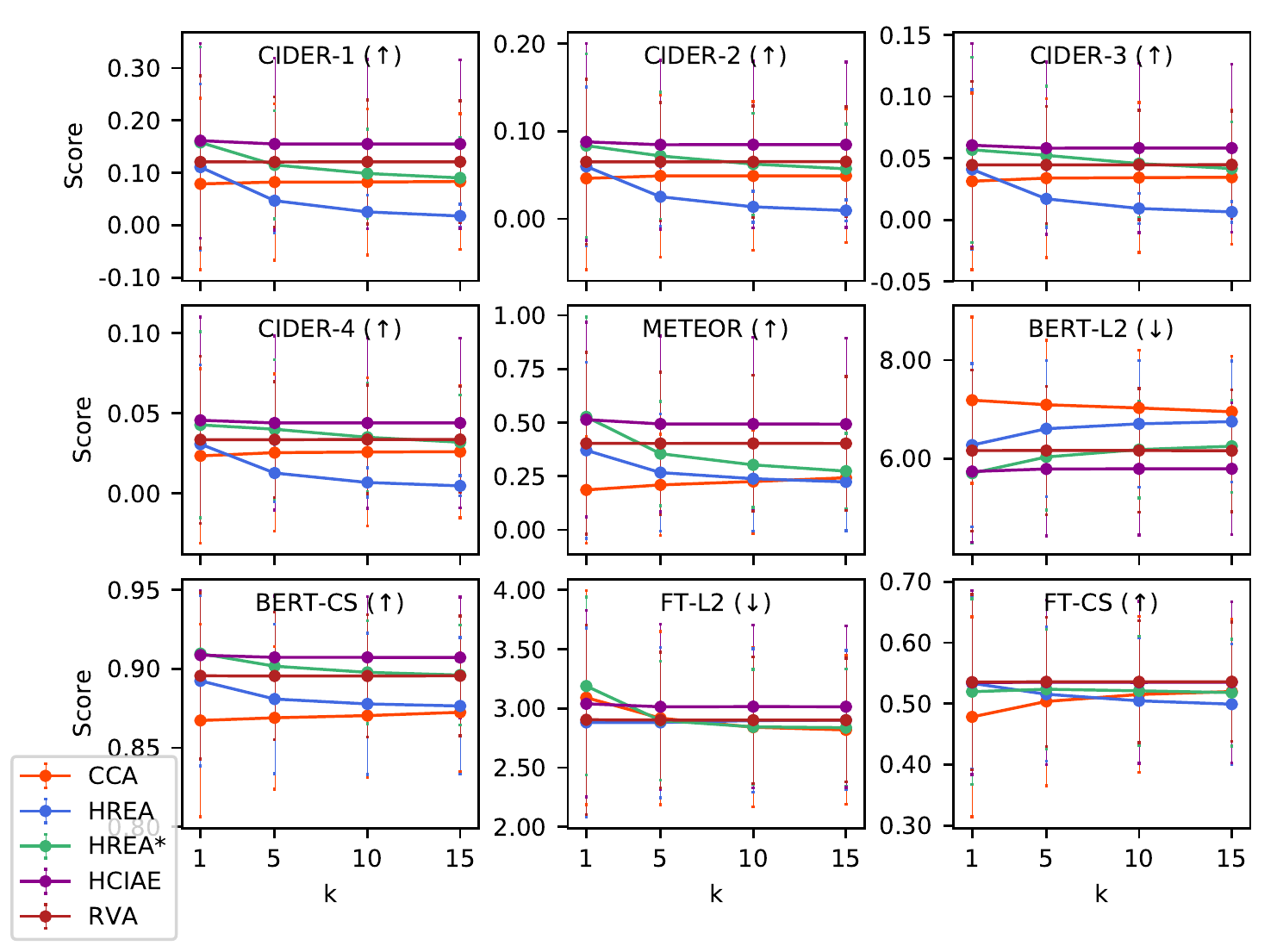}
\vspace*{-0.8em}
\caption{%
  Overlap and embedding distance metrics for \(k=1,5,10,15\) generations from \gls{SOTA} models on \(\mathcal{H}_v\) (top) and the full validation set (bottom).
\gls{CCA}\textsc{-aq-g} and \textsc{rva} generally show improving trends with increasing \(k\), which isn't the case for \textsc{hrea-qih-g} variants.
}
\label{fig:ksample}
% \vspace*{-2em}
\end{figure}

Beyond just \(k=1\) generation, a particularly useful feature of our revised scheme is that, unlike the rank-based evaluation, it can evaluate across any \(k\) number of generations sampled from the models (see~\cref{fig:ksample}).
Answer correctness can therefore be measured, without penalising diversity, even if the generations fall outside the candidate set for the given question.
This yields an interesting insight: for some models (notably the \textsc{hrea-qih-g} variants) performance degrades as \(k\) increases---a useful thing to know if deploying this model as a \gls{VD} agent in the real-world!
Others, like \textsc{rva}, \textsc{hciae-g-dis} and \gls{CCA}\textsc{-aq-g}, generally remain constant or improve with higher \(k\).
Surprisingly, \gls{CCA}\textsc{-aq-g}, despite its poorer absolute performance across the metrics at \(k=1\), holds its own and even improves with increasing \(k\).
This allows us to compare models' generation capabilities and indeed robustness in the answering task---something not possible with the rank-based evaluation.

\section{Discussion}
\label{sec:discussion}

In this paper, we propose a revised evaluation suite for \textit{VisDial} drawing on existing metrics from the \textsc{nlp} community that measures similarity between answers generated by a model and a given reference set of answers.
We arrive at the need for alternate evaluations through the findings of~\citet{massiceti2018novd} and our own analysis of existing evaluation metrics on the \textit{VisDial} dataset, which we show can suffer from a number of issues to do with a mismatch between the \gls{VD} task and an evaluation for it that depends on ranking metrics.
While a recent update to the evaluation paradigm of \textit{VisDial} incorporates both human judgements of answer validity and multiple plausible answers into a final score, issues relating to ranking persist, albeit to a lesser extent.
Here, we advocate use of answers directly \emph{generated} by a model, in concert with consensus-based metrics measuring similarity against \emph{sets} of answers marked as valid by human annotators.

It is practically infeasible to obtain these validity judgements at scale, however, thus restricting the extent to which the revised scheme can be applied.
To address this issue, we develop a semi-supervised automated mechanism to extract sets of relevant answers from given candidate sets, using sparse human annotations and correlations through \gls{CCA}.
We verify these sets by computing their intersection with those marked by humans, asking turkers via \gls{AMT}, and measuring their utility for the \gls{VD} task.
Based on such experiments, we expand the \textit{VisDial} dataset with these reference set annotations and release this and the revised evaluation scheme as \newdataset{} for future evaluation and model development.
We intend this to be one possible improvement in the face of inherent constraints on the \textit{VisDial} dataset, and hope that the community adopts the revised evaluation going forwards.

{
\vspace*{0.5ex}
\small 
\setstretch{1.2}
\noindent \textbf{Acknowledgments}
This work was supported by ERC grant ERC-2012-AdG 321162-HELIOS, EPSRC grant Seebibyte EP/M013774/1, EPSRC/MURI grant EP/N019474/1, FAIR \emph{ParlAI} grant, the Skye Foundation and the Toyota Research Institute.
We thank Yulei Niu for his help with the \textsc{rva} code.\par}

\bibliography{references}
\bibliographystyle{natbib}

\newpage
\clearpage
\appendix

\section{Detailed \acrshort{CCA} results}
\label{sec:app-cca-analysis}
We extend the work in~\citet{massiceti2018novd} and conduct a detailed rank-based performance analysis of \gls{CCA} versus \gls{SOTA} approaches across both \textit{v0.9} and \textit{v1.0} of the \textit{VisDial} dataset (see~\cref{tab:full-cca-analysis}).

\paragraph{Feature extractors}
We represent the words in the questions/answers as 300-dimensional \textit{FastText}~\citep{bojanowski2017enriching} embeddings.
To obtain (300-dimensional) sentence representations, we average word embeddings, following generally received intuition~\citep{aurora2017simple,wieting2017random}, padding or truncating up to 16 words following~\citet{massiceti2018novd}.
While not relevant in the main paper, here we represent images as 512-dimensional pre-trained \textit{ResNet34}~\citep{He_2016deep} features, taking the output of \emph{conv5}, after the \emph{avg pool}.
Our choice of feature extractors is largely arbitrary, however, so to rule out gains from one feature extractor over another and to compare against \gls{SOTA} models~\citep{lu2017best,wu2017you, Das_VisualDialog} which use others, we also employ \textit{GloVe}~\citep{pennington2014glove} (300-dimensional, from Common Crawl-42B tokens), and \textit{VGG-16/19}~\citep{Simonyan_2014c} (4096-dimensional, extracted after the second \textit{fc-4096} layer) features for the questions/answers and images, respectively.

\paragraph{Baselines}

We compare against ablative versions (without image and/or dialogue history) of the \gls{SOTA} models, as well as two nearest-neighbour baselines, as established in~\citep{Das_VisualDialog}:
\begin{compactdesc}
\item[\textsc{nn-aq}:] given a test question, we find the \(k\) nearest-neighbour questions (by average {\it GloVe} embedding) from the training set.
  We take the mean of their \(k\) corresponding answers (again in GloVe embedding) to represent a ``canonical'' answer to that question, ranking the test question's candidate answers by their \(L_2\) distances to it.
\item[\textsc{nn-aqi}:] given a test question and image, we first draw the \(k\) nearest-neighbour questions to the test question from the training set.
  From this set, we draw the \(k'\) questions whose corresponding image features are most similar to the test image feature.
  Taking the mean of their \(k'\) corresponding answers, we then rank the test question's candidate answers as above (\(k=100\), \(k'=20\) as per~\citep{Das_VisualDialog}).
\end{compactdesc}
We also include \textsc{ncca-aq} (no-\gls{CCA}), by computing correlation, as centred cosine distance, directly between the features for questions and answers.
Note that for \gls{CCA}\textsc{-aqi (q)}, at test time, correlation is computed between questions and candidate answers \(\bm{A}\) using projection matrices learned using images (I) as well.
Note too that \acrshort{NDCG} scores are not computed for \textit{v0.9} since human annotations on answer relevance are not available.

Similar to~\citet{massiceti2018novd}, we observe that the \gls{MR} achieved by \gls{CCA} is similar to that of \gls{SOTA} models, despite the approach's light-weight nature.
Comparing to the nearest-neighbour baselines, \gls{CCA} is superior in \gls{MR}, and additionally in computation and storage requirements since a nearest-neighbour approach requires the train data (including images) at test time.

\begin{table}[!ht]
  \centering \footnotesize
  \newcommand{\rt}[2]{\parbox[t]{2mm}{\multirow{#1}{*}{\rotatebox[origin=c]{90}{#2}}}}
  \newcommand{\citelink}[2]{#2 \citep{#1}}
  \def\hci{\textsc{hciae-g-dis}}
  \def\coa{\textsc{CoAtt-GAN}}
  \def\hrea{\textsc{hrea-qih-g}}
  \def\hre{\textsc{hre-qih-g}}
  \def\hreQH{\textsc{hre-qh-g}}
  \def\hreQ{\textsc{hre-q-g}}
  \def\lfQ{\textsc{lf-q-g}}
  \def\lfQH{\textsc{lf-qh-g}}
  \def\lfQI{\textsc{lf-qi-g}}
  \def\lfQIH{\textsc{lf-qih-g}}
  \def\mn{\textsc{mn-qih-g}}
  \def\mnatt{\textsc{mna-qih-g}}
  \def\rva{\textsc{rva}}
  \def\ditto{\multicolumn{1}{c}{''}}
  \def\s{\hspace*{5pt}}
  \scalebox{0.65}{%
  \begin{tabular}{@{}c@{\s}l@{\s}>{\footnotesize}c@{\s}l@{\s}c@{\s}c@{\s}c@{\s}c@{\s}c@{\s}c@{}}
    \toprule
    &
    & Model                     & I/QA features         & MR          & R@1   & R@5   & R@10  & MRR    & NDCG   \\
    \midrule
    \rt{13}{\acrshort{SOTA}}
    & \rt{7}{\emph{v0.9}}
    & \hci                      & VGG-19/learned        & {\bf 14.23} & 44.35 & 65.28 & 71.55 & 0.5467 & -      \\
    &
    & \coa                      & VGG-16/learned        & {\bf 14.43} & 46.10 & 65.69 & 71.74 & 0.5578 & -      \\
    &
    & \hrea                     & \ditto                & {\bf 16.60} & 42.13 & 62.44 & 68.42 & 0.5238 & -      \\
    &
    & \lfQIH                    & \ditto                & {\bf 16.76} & 40.86 & 62.05 & 68.28 & 0.5146 & -      \\
    &
    & \hre                      & \ditto                & {\bf 16.97} & 42.23 & 62.28 & 68.11 & 0.5237 & -      \\
    % &
    % & \hreGH                    & \ditto                & 17.36       & 40.00 & 61.54 & 67.39 & 0.5089 & -      \\
    &
    & \lfQI                     & \ditto                & 17.06       & 42.06 & 61.65 & 67.60 & 0.5206 & -      \\
    % &
    % & \lfQH                     & \ditto                & 17.40       & 40.12 & 61.55 & 67.40 & 0.5099 & -      \\
    &
    & \lfQ                      & \ditto                & 17.80       & 39.74 & 60.67 & 66.49 & 0.5048 & -      \\
    \cmidrule(l){2-10}
    & \rt{7}{\emph{v1.0}}
    & \hre                      & VGG-16/learned        & 18.78       & 34.78 & 56.18 & 63.72 & 0.4561 & 0.5245 \\
    &
    & \lfQIH                    & \ditto                & 18.81       & 35.08 & 55.92 & 64.02 & 0.4568 & 0.5121 \\
    &
    & \hrea                     & \ditto                & 19.15       & 34.73 & 56.55 & 63.18 & 0.4555 & 0.5189 \\
    &
    & \mn                       & \ditto                & 21.25       & 33.83 & 54.03 & 60.65 & 0.4415 & 0.5074 \\
    &
    & \mnatt                    & \ditto                & 21.14       & 34.55 & 53.57 & 60.08 & 0.4451 & 0.5084 \\
    &
    & \rva       & ResNet101 + {\sc rpn}/learned                          & 18.50       & 38.73 & 57.70 & 64.73 & 0.4827 & 0.5574 \\
    \midrule
    \rt{6}{Baselines}
    & \rt{6}{\emph{v0.9}}
    & \multirow{2}*{\textsc{nn-aq}}     & GloVe                 & 19.67       & 29.88 & 47.07 & 55.44 & 0.3898 & -      \\
    &
    &                           & FastText              & 25.92       & 19.86 & 34.74 & 43.55 & 0.2830 & -      \\
    \cmidrule(l){3-10}
    &
    & \multirow{2}*{\textsc{nn-aqi}}    & VGG-16/GloVe          & 20.14       & 29.93 & 46.42 & 54.76 & 0.3873 & -      \\
    &
    &                           & ResNet34/FastText     & 25.88       & 21.19 & 35.78 & 44.31 & 0.2941 & -      \\
    \cmidrule(l){3-10}
    &
    & \textsc{ncca-aq}                    & FastText              & 57.18       & 4.13  & 9.67 & 13.89 & 0.0837 & -      \\
    \midrule
    \rt{10}{\acrshort{CCA}}
    & \rt{5}{\emph{v0.9}}
    & \multirow{2}*{\textsc{aq}}        & GloVe                 & {\bf 15.86} & 16.93 & 44.83 & 58.44 & 0.3044 & -      \\
    &
    &                           & FastText              & {\bf 16.21} & 16.85 & 44.96 & 58.10 & 0.3041 & -      \\
    \cmidrule(l){3-10}
    &
    & \multirow{3}*{\textsc{aqi(q)}}   & ResNet34/FastText     & 18.27       & 12.24 & 35.55 & 50.88 & 0.2439 & -      \\
    &
    &                           & VGG-16/GloVe          & 26.03       & 12.24 & 30.96 & 42.63 & 0.2237 & -      \\
    &
    &                           & VGG-19/GloVe          & 18.88       & 12.42 & 34.52 & 48.47 & 0.2409 & -      \\
    \cmidrule(l){2-10}
    & \rt{5}{\emph{v1.0}}
    & \multirow{2}*{\textsc{aq}}        & GloVe                 & {\bf 16.60} & 16.10 & 39.38 & 54.68 & 0.2824 & 0.3504 \\
    &
    &                           & FastText              & {\bf 17.07} & 16.18 & 40.18 & 55.35 & 0.2845 & 0.3493 \\
    \cmidrule(l){3-10}
    &
    & \multirow{3}*{\textsc{aqi (q)}}   & ResNet34/FastText     & 19.25       & 12.63 & 32.88 & 48.68 & 0.2379 & 0.3077  \\
    &
    &                           & VGG-16/GloVe          & 19.11       & 13.53 & 32.43 & 47.13 & 0.2415 & 0.3071  \\
    &
    &                           & VGG-19/GloVe          & 19.29       & 13.38 & 32.73 & 47.23 & 0.2415 & 0.3000  \\
    \bottomrule
  \end{tabular}}
  \vspace*{-1ex}
  \caption{%
    Results for \gls{SOTA} vs.\ \gls{CCA} on the \textit{VisDial} \textit{v0.9} and \textit{v1.0} dataset.
    \gls{CCA} achieves comparable performance in mean rank (MR) while ignoring both image and dialogue sequence.
    Note, {\sc rpn}~\citep{renNIPS15fasterrcnn}.
  }
  \label{tab:full-cca-analysis}
  \vspace*{1ex}
\end{table}

% \vspace{-0.5ex}
\section{\acrshort{NDCG} details}
\label{sec:app-ndcg}

The \gls{NDCG} is the ratio of the \gls{DCG} of a model's predicted ranking to the \gls{DCG} of the `ideal' ranking, obtained by sorting the relevance scores in descending order as \(\acrshort{NDCG}@m = \frac{\gls{DCG}@m}{\text{ideal} \: \gls{DCG}@m}\),
% %
% \begin{align*}
%   \acrshort{NDCG}@m = \frac{\gls{DCG}@m}{\text{ideal} \: \gls{DCG}@m}
% \end{align*}
%
%\vspace*{-1em}
where \(m\) is the number of answers with human-derived relevance scores in the set of 100, and \(\gls{DCG}@m = \sum_i^m \frac{\rho_i}{\log_2(i+1)}\).
% %
% \begin{align*}
%   \gls{DCG}@m = \sum_i^m \frac{\rho_i}{\log_2(i+1)}
% \end{align*}
%
where \(i\) is the rank of the answer candidate, and \(\rho_i\) is the human-assigned relevance score of the \(i\)-th ranked answer.

\section{Full revised evaluation with \(H\) and \(\Sigma\)}
\label{sec:app-consensus}

\begin{table*}[!p]
\centering
\def\s{\hspace*{3ex}}
\def\j{0.8ex}
\def\mc#1#2{\multicolumn{#1}{c}{{\sc #2}}}
\def\up{\(\uparrow\)}
\def\down{\(\downarrow\)}
\scalebox{0.58}{
\setlength{\tabcolsep}{3pt}
\begin{tabular}{@{}l*{4}{c}@{\s}*{4}{c}@{\s}c@{\s}*{2}{c}@{\s}*{2}{c}@{}}
\toprule
Model                  & \mc{4}{cider\up}                  & \mc{4}{bleu\up}                   & {\sc meteor\up} & \mc{2}{bert}    & \mc{2}{FastText} \\
                         \cmidrule(lr){2-5}                \cmidrule(lr){6-9}                  \cmidrule(lr){11-12}                \cmidrule(lr){13-14}
                         & n=1    & n=2    & n=3    & n=4    & n=1    & n=2    & n=3    & n=4    &              & \(L_2\)\down& CS\up     & \(L_2\)\down& CS\up      \\
\midrule
\(\gtA\)  \hfill \(\mu\)        &  0.1889 & 0.1253 & 0.0986 & 0.0819 & 0.9961 & 0.4840 & 0.3934 & 0.2916 &  0.9971 & 5.6087 &  0.9072 &  2.6576 &  0.5681  \\
%\hfill \(\sigma\)               & (0.1622) & (0.1126) & (0.0955) & (0.0863) & (0.0382) & (0.4969) & (0.4859) & (0.4503) & (0.0405) & (1.3892) &  (0.0421) &  (0.7489) &  (0.1178) \\
\(\Gamma_H\) \hfill \(\mu\)     & 0.1915 & 0.1437 & 0.1165 & 0.0962 & 1.0000 & 0.7633 & 0.5974 & 0.3371 & 1.0000 & 5.4297 &  0.9120 &  2.3339 &  0.6067 \\
\hfill \(\sigma\)               & (0.1490) & (0.1051) & (0.0869) & (0.0772)& (0.0000) & (0.1440) & (0.1538) & (0.1675) & (0.0000) & (1.0121) &  (0.0254) &  (0.4738) &  (0.0850)  \\
\hfill \(\gamma\)               & 0.2765 & 0.2151 & 0.1810 & 0.1513 & 1.0000 & 0.9898 & 0.9826 & 0.9300 & 1.0000 & 4.7000 &  0.9334 &  1.8757 &  0.6992 \\
\cmidrule{2-14}
\gls{CCA}\textsc{-aq-g} (k=1)    \hfill \(\mu\)      & 0.0721 & 0.0434 & 0.0298 & 0.0226 & 0.4323 & 0.1461 & 0.0345 & 0.0138 & 0.2713 & 7.1231 &  0.8690 &  3.1251 &  0.4555 \\
\hfill \(\sigma\)                     & (0.1187) & (0.0726) & (0.0506) & (0.0389) & (0.3839) & (0.3189) & (0.1594) & (0.1041) & (0.3083) & (1.7118) &  (0.0626) &  (0.8266) &  (0.1364) \\[\j]
% \hfill \(\gamma\)                     & 0.0721 & 0.0434 & 0.0298 & 0.0226 & 0.4323 & 0.1461 & 0.0345 & 0.0138 & 0.2713 & 7.1231 &  0.8690 &  3.1251 &  0.4555 \\[\j]
%
\gls{CCA}\textsc{-aq-g} (k=5) \hfill \(\mu\)   & 0.0719 & 0.0431 & 0.0298 & 0.0225 & 0.4195 & 0.1569 & 0.0451 & 0.0131 & 0.2712 & 7.1000 &  0.8685 &  2.9570 &  0.4787 \\
\hfill \(\sigma\)               &(0.1094) & (0.0640) & (0.0443) & (0.0337) & (0.3169) & (0.2559) & (0.1312) & (0.0592) & (0.2613) & (1.2868) &  (0.0455) &  (0.6648) &  (0.1150) \\
\hfill \(\gamma\)               & 0.1108 & 0.0690 & 0.0488 & 0.0373 & 0.6330 & 0.3272 & 0.1275 & 0.0478 & 0.4162 & 6.0801 &  0.9031 &  2.5311 &  0.5571 \\[\j]
\gls{CCA}\textsc{-aq-g} (k=10) \hfill \(\mu\)  & 0.0708 & 0.0420 & 0.0291 & 0.0221 & 0.4156 & 0.1539 & 0.0516 & 0.0149 & 0.2721 & 7.0301 &  0.8702 &  2.8850 &  0.4889 \\
\hfill \(\sigma\)               &(0.1071) & (0.0603) & (0.0418) & (0.0317) & (0.2931) & (0.2183) & (0.1248) & (0.0500) & (0.2410) & (1.1316) &  (0.0393) &  (0.6072) &  (0.1077) \\
\hfill \(\gamma\)               & 0.1320 & 0.0840 & 0.0609 & 0.0470 & 0.7251 & 0.4445 & 0.2276 & 0.0960 & 0.5047 & 5.7117 &  0.9132 &  2.3446 &  0.5984 \\[\j]
\gls{CCA}\textsc{-aq-g} (k=15) \hfill \(\mu\)   & 0.0715 & 0.0419 & 0.0293 & 0.0222 & 0.4230 & 0.1499 & 0.0585 & 0.0146 & 0.2859 & 6.9386 &  0.8728 &  2.8662 &  0.4935\\
\hfill \(\sigma\)                & (0.1061) & (0.0583) & (0.0402) & (0.0304) & (0.2828) & (0.1873) & (0.1160) & (0.0441) & (0.2375) & (1.0861) &  (0.0372) &  (0.5732) &  (0.1028)\\
\hfill \(\gamma\)               & 0.1498 & 0.0957 & 0.0704 & 0.0545 & 0.7858 & 0.5191 & 0.3085 & 0.1237 & 0.5908 & 5.5281 &  0.9178 &  2.2542 &  0.6191\\
%
%\gls{CCA}\textsc{-iqa-g} \hfill \(\mu\)        &  0.0649 & 0.0400 & 0.0277 & 0.0211 & 0.4107 & 0.1407 & 0.0348 & 0.0153 & 0.2581 &  7.2405 &  0.8641  &  3.1807 &  0.4411 \\

%\hfill \(\sigma\)               & (0.1090) & (0.0701) & (0.0496) & (0.0384) & (0.3799) & (0.3130) & (0.1627) & (0.1132) & (0.3052) & (1.7973) &  (0.0666) &  (0.8510) &  (0.1322) \\
\cmidrule{2-14}
\textsc{hrea-qih-g} (k=1)  \hfill \(\mu\)     & 0.0880 & 0.0483 & 0.0333 & 0.0252 & 0.5557 & 0.0948 & 0.0411 & 0.0136 & 0.4813 & 6.2875 &  0.8927 &  2.9724 &  0.5079  \\
\hfill \(\sigma\)               & (0.1161) & (0.0671) & (0.0480) & (0.0374) & (0.4252) & (0.2451) & (0.1799) & (0.1082) & (0.4264) & (1.5281) &  (0.0536) &  (0.7308) &  (0.1146)  \\[\j]
\textsc{hrea-qih-g} (k=5)  \hfill \(\mu\)      & 0.0371 & 0.0202 & 0.0137 & 0.0104 & 0.4343 & 0.0741 & 0.0204 & 0.0048 & 0.3353 & 6.5871 &  0.8820 &  2.9495 &  0.4920 \\
\hfill \(\sigma\)               & (0.0493) & (0.0269) & (0.0184) & (0.0140) & (0.2975) & (0.1363) & (0.0777) & (0.0400) & (0.2768) & (1.2200) &  (0.0450) &  (0.5736) &  (0.0896) \\
\hfill \(\gamma\)               & 0.1171 & 0.0644 & 0.0442 & 0.0335 & 0.6983 & 0.1877 & 0.0596 & 0.0154 & 0.5821 & 5.6691 &  0.9127 &  2.5382 &  0.5625 \\[\j]
\textsc{hrea-qih-g} (k=10) \hfill \(\mu\) & 0.0558 & 0.0303 & 0.0206 & 0.0156 & 0.3919 & 0.0648 & 0.0169 & 0.0039 & 0.2958 & 6.6661 &  0.8796 &  2.9599 &  0.4907  \\
\hfill \(\sigma\)               &(0.0716) & (0.0405) & (0.0290) & (0.0232) & (0.2691) & (0.1190) & (0.0663) & (0.0346) & (0.2452) & (1.1225) &  (0.0415) &  (0.5827) &  (0.0906) \\
\hfill \(\gamma\)               & 0.1194 & 0.0656 & 0.0450 & 0.0341 & 0.7154 & 0.1975 & 0.0607 & 0.0154 & 0.5915 & 5.6003 &  0.9149 &  2.5639 &  0.5600 \\[\j]
\textsc{hrea-qih-g} (k=15) \hfill \(\mu\)& 0.0137 & 0.0074 & 0.0051 & 0.0038 & 0.3704 & 0.0606 & 0.0156 & 0.0037 & 0.2767 & 6.6924 &  0.8788 &  2.9571 &  0.4776 \\
\hfill \(\sigma\)               & (0.0178) & (0.0096) & (0.0066) & (0.0050) & (0.2533) & (0.1107) & (0.0619) & (0.0334) & (0.2278) & (1.0705) &  (0.0397) &  (0.5238) &  (0.0809) \\
\hfill \(\gamma\)               & 0.1207 & 0.0663 & 0.0455 & 0.0344 & 0.7227 & 0.2014 & 0.0609 & 0.0154 & 0.5959 & 5.5717 &  0.9158 &  2.4382 &  0.5727 \\
\cmidrule{2-14}
\textsc{hrea-qih-g*} (k=1) \hfill \(\mu\)       & 0.1359 & 0.0721 & 0.0494 & 0.0372 & 0.7646 & 0.0614 & 0.0374 & 0.0064 & 0.7149 & 5.5727 &  0.9149  &  3.2664 &  0.4971 \\
\hfill \(\sigma\)               & (0.1431) & (0.0789) & (0.0564) & (0.0430) & (0.4109) & (0.2279) & (0.1875) & (0.0711) & (0.4291) & (1.0037) &  (0.0278)  &  (0.6524) &  (0.1148) \\[\j]
\textsc{hrea-qih-g*} (k=5) \hfill \(\mu\) & 0.1054 & 0.0660 & 0.0480 & 0.0368 & 0.6282 & 0.2607 & 0.1416 & 0.0382 & 0.5027 & 5.9013 &  0.9063 &  2.9186 &  0.5127 \\
\hfill \(\sigma\)                 & (0.0791) & (0.0498) & (0.0367) & (0.0282) & (0.2451) & (0.2145) & (0.1688) & (0.0874) & (0.2288) & (0.7756) &  (0.0253) &  (0.4315) &  (0.0808) \\
\hfill \(\gamma\)                & 0.2222 & 0.1472 & 0.1119 & 0.0869 & 0.9582 & 0.6808 & 0.4846 & 0.1623 & 0.9268 & 5.0449 &  0.9274 &  2.2062 &  0.6328 \\[\j]
\textsc{hrea-qih-g*} (k=10) \hfill \(\mu\) & 0.0937 & 0.0600 & 0.0438 & 0.0337 & 0.5825 & 0.2739 & 0.1423 & 0.0461 & 0.4310 & 6.0202 &  0.9031 &  2.8534 &  0.5117 \\
\hfill \(\sigma\)               &(0.0656) & (0.0409) & (0.0300) & (0.0231) & (0.2271) & (0.1959) & (0.1412) & (0.0808) & (0.1868) & (0.7250) &  (0.0241) &  (0.4241) &  (0.0761) \\
\hfill \(\gamma\)               & 0.2456 & 0.1701 & 0.1326 & 0.1042 & 0.9780 & 0.8198 & 0.6472 & 0.2946 & 0.9562 & 4.9441 &  0.9295 &  2.0886 &  0.6584\\[\j]
\textsc{hrea-qih-g*} (k=15) \hfill \(\mu\) & 0.0852 & 0.0546 & 0.0398 & 0.0307 & 0.5474 & 0.2635 & 0.1321 & 0.0444 & 0.3866 & 6.0859 &  0.9014 &  2.8411 &  0.5094 \\
\hfill \(\sigma\)               & (0.0591) & (0.0362) & (0.0263) & (0.0202) & (0.2174) & (0.1834) & (0.1265) & (0.0721) & (0.1651) & (0.7063) &  (0.0237) &  (0.4329) &  (0.0750) \\
\hfill \(\gamma\)               & 0.2537 & 0.1793 & 0.1410 & 0.1113 & 0.9815 & 0.8638 & 0.7045 & 0.3525 & 0.9653 & 4.8995 &  0.9305 &  2.0457 &  0.6686 \\
\cmidrule{2-14}

\textsc{mna-qih-g*} (k=1) \hfill \(\mu\)    & 0.1365 & 0.0713 & 0.0484 & 0.0364 & 0.7713 & 0.0360 & 0.0225 & 0.0019 & 0.7316 & 5.5549 &  0.9153 &  3.3271 &  0.4920 \\
\hfill \(\sigma\)                &(0.1394) & (0.0767) & (0.0540) & (0.0405) & (0.4134) & (0.1796) & (0.1467) & (0.0333) & (0.4274) & (1.0043) &  (0.0279) &  (0.6399) &  (0.1153) \\[\j]

\textsc{mna-qih-g*} (k=5) \hfill \(\mu\)    & 0.1066 & 0.0666 & 0.0485 & 0.0370 & 0.6397 & 0.2592 & 0.1485 & 0.0357 & 0.5189 & 5.8719 &  0.9070 &  2.9412 &  0.5104 \\
\hfill \(\sigma\)                &(0.0743) & (0.0464) & (0.0344) & (0.0263) & (0.2460) & (0.2148) & (0.1749) & (0.0855) & (0.2298) & (0.7922) &  (0.0260) &  (0.4207) &  (0.0781) \\
\hfill \(\gamma\)               & 0.2276 & 0.1495 & 0.1137 & 0.0876 & 0.9676 & 0.6748 & 0.4981 & 0.1489 & 0.9425 & 5.0193 &  0.9277 &  2.2086 &  0.6328 \\[\j]

\textsc{mna-qih-g*} (k=10) \hfill \(\mu\)     & 0.0941 & 0.0604 & 0.0443 & 0.0341 & 0.5955 & 0.2786 & 0.1481 & 0.0479 & 0.4483 & 5.9945 &  0.9037 &  2.8828 &  0.5097 \\
\hfill \(\sigma\)                &(0.0598) & (0.0377) & (0.0277) & (0.0214) & (0.2314) & (0.1932) & (0.1398) & (0.0811) & (0.1969) & (0.7360) &  (0.0247) &  (0.4223) &  (0.0744) \\
\hfill \(\gamma\)               & 0.2488 & 0.1731 & 0.1358 & 0.1065 & 0.9805 & 0.8228 & 0.6819 & 0.2986 & 0.9640 & 4.9167 &  0.9298 &  2.0722 &  0.6621 \\[\j]

\textsc{mna-qih-g*} (k=15) \hfill \(\mu\)    & 0.0854 & 0.0551 & 0.0404 & 0.0311 & 0.5590 & 0.2709 & 0.1386 & 0.0472 & 0.4028 &  6.0637 &  0.9019 &  2.8721 &  0.5076 \\
\hfill \(\sigma\)                & (0.0535) & (0.0333) & (0.0244) & (0.0189) & (0.2206) & (0.1808) & (0.1239) & (0.0740) & (0.1743) & (0.7067) &  (0.0239) &  (0.4318) &  (0.0735) \\
\hfill \(\gamma\)               & 0.2585 & 0.1835 & 0.1453 & 0.1147 & 0.9856 & 0.8770 & 0.7516 & 0.3688 & 0.9733 & 4.8716 &  0.9308 &  2.0301 &  0.6718 \\
\cmidrule{2-14}

\textsc{hciae-g-dis} (k=1) \hfill \(\mu\) & 0.1338 & 0.0718 & 0.0493 & 0.0372 & 0.7515 & 0.0945 & 0.0513 & 0.0104 & 0.6758 & 5.6690 &  0.9122 &  3.1551 &  0.5049 \\
\hfill \(\sigma\) & (0.1478) & (0.0809) & (0.0564) & (0.0426) & (0.4009) & (0.2627) & (0.2092) & (0.0926) & (0.4291) & (1.0671) &  (0.0304) &  (0.6940) &  (0.1158) \\
\hfill \(\gamma\) & 0.1338 & 0.0718 & 0.0493 & 0.0372 & 0.7515 & 0.0945 & 0.0513 & 0.0104 & 0.6758 & 5.6690 &  0.9122 &  3.1551 &  0.5049 \\[\j]

\textsc{hciae-g-dis} (k=5) \hfill \(\mu\) & 0.1278 & 0.0689 & 0.0473 & 0.0357 & 0.7280 & 0.1015 & 0.0518 & 0.0121 & 0.6501 & 5.7489 &  0.9099 &  3.1158 &  0.5062 \\
\hfill \(\sigma\) & (0.1274) & (0.0672) & (0.0464) & (0.0350) & (0.3462) & (0.2129) & (0.1559) & (0.0701) & (0.3738) & (0.9951) &  (0.0287) &  (0.6097) &  (0.1028) \\
\hfill \(\gamma\) & 0.1685 & 0.0944 & 0.0662 & 0.0502 & 0.8672 & 0.2178 & 0.1282 & 0.0348 & 0.7995 & 5.4049 &  0.9192 &  2.8464 &  0.5460 \\[\j]

\textsc{hciae-g-dis} (k=10) \hfill \(\mu\) & 0.1275 & 0.0689 & 0.0473 & 0.0357 & 0.7290 & 0.1050 & 0.0522 & 0.0122 & 0.6489 & 5.7448 &  0.9101 &  3.1146 &  0.5068 \\
\hfill \(\sigma\) & (0.1259) & (0.0665) & (0.0459) & (0.0347) & (0.3403) & (0.2082) & (0.1492) & (0.0654) & (0.3678) & (0.9929) &  (0.0286) &  (0.6018) &  (0.1014) \\
\hfill \(\gamma\) & 0.1782 & 0.1019 & 0.0721 & 0.0549 & 0.8917 & 0.2774 & 0.1653 & 0.0525 & 0.8335 & 5.3337 &  0.9212 &  2.7530 &  0.5616 \\[\j]

\textsc{hciae-g-dis} (k=15) \hfill \(\mu\) & 0.1270 & 0.0683 & 0.0469 & 0.0354 & 0.7266 & 0.1015 & 0.0516 & 0.0121 & 0.6464 & 5.7557 &  0.9097 &  3.1179 &  0.5065 \\
\hfill \(\sigma\) & (0.1245) & (0.0655) & (0.0451) & (0.0340) & (0.3374) & (0.2010) & (0.1448) & (0.0637) & (0.3635) & (0.9823) &  (0.0285) &  (0.5934) &  (0.1007) \\
\hfill \(\gamma\) & 0.1848 & 0.1069 & 0.0763 & 0.0581 & 0.9113 & 0.3150 & 0.2019 & 0.0644 & 0.8556 & 5.2762 &  0.9225 &  2.6875 &  0.5707 \\

\cmidrule{2-14}
\textsc{rva} (k=1) \hfill \(\mu\)  & 0.1042 & 0.0563 & 0.0385 & 0.0291 & 0.6135 & 0.1001 & 0.0433 & 0.0119 & 0.5328 & 6.1466 &  0.8967 &  2.9543 &  0.5161 \\
\hfill \(\sigma\)     &(0.1311) & (0.0709) & (0.0491) & (0.0371) & (0.4193) & (0.2532) & (0.1884) & (0.0993) & (0.4304) & (1.4871) &  (0.0509) &  (0.7253) &  (0.1170) \\[\j]
%\hfill \(\gamma\)   & 0.1042 & 0.0563 & 0.0385 & 0.0291 & 0.6135 & 0.1001 & 0.0433 & 0.0119 & 0.5328 &  6.1466 &  0.8967 &  2.9543 &  0.5161 \\
%
\textsc{rva} (k=5) \hfill \(\mu\)  & 0.1038 & 0.0560 & 0.0383 & 0.0289 & 0.6088 & 0.0983 & 0.0435 & 0.0112 & 0.5319 & 6.1182 &  0.8977 &  2.9773 &  0.5139 \\
\hfill \(\sigma\)     &(0.1038) & (0.0542) & (0.0370) & (0.0279) & (0.2939) & (0.1435) & (0.0969) & (0.0453) & (0.3045) & (1.0522) &  (0.0338) &  (0.5143) &  (0.0873) \\
\hfill \(\gamma\)   & 0.1777 & 0.1008 & 0.0710 & 0.0540 & 0.8968 & 0.3193 & 0.1727 & 0.0510 & 0.8315 &   5.3013 &  0.9219 &  2.4047 &  0.600\\[\j]
\textsc{rva} (k=10) \hfill \(\mu\)   & 0.1025 & 0.0553 & 0.0379 & 0.0286 & 0.6118 & 0.0997 & 0.0416 & 0.0114 & 0.5333 & 6.1314 &  0.8974 &  2.9757 &  0.5129 \\
\hfill \(\sigma\)     &(0.0964) & (0.0503) & (0.0344) & (0.0259) & (0.2729) & (0.1265) & (0.0815) & (0.0380) & (0.2869) & (0.9821) &  (0.0314) &  (0.4743) &  (0.0803) \\
\hfill \(\gamma\)   & 0.1995 & 0.1180 & 0.0845 & 0.0646 & 0.9450 & 0.4660 & 0.2625 & 0.0887 & 0.8963 & 5.1621 &  0.9253 &  2.2568 &  0.6254 \\[\j]
\textsc{rva} (k=15) \hfill \(\mu\)   & 0.1034 & 0.0559 & 0.0383 & 0.0289 & 0.6147 & 0.1011 & 0.0427 & 0.0124 & 0.5366 & 6.1234 &  0.8976 &  2.9739 &  0.5138 \\
\hfill \(\sigma\)     & (0.0946) & (0.0492) & (0.0336) & (0.0253) & (0.2652) & (0.1208) & (0.0784) & (0.0383) & (0.2781) & (0.9575) &  (0.0305) &  (0.4630) &  (0.0788) \\
\hfill \(\gamma\)   & 0.2110 & 0.1290 & 0.0936 & 0.0722 & 0.9575 & 0.5597 & 0.3335 & 0.1249 & 0.9209 & 5.0983 &  0.9267 &  2.1786 &  0.6389 \\
\bottomrule
\end{tabular}}
\vspace*{-1ex}
{\caption{
  Overlap and embedding distance metrics computed for \(k\) generations against the human-annotated reference set \(H\) on the validation subset \(\mathcal{H}_v\).
  For \textsc{hrea-qih-g}, on average \(\sim\)6 answers are the empty string, which are excluded from the computation.
  Metrics marked~\(\uparrow\) indicate higher values are better, and those marked~\(\downarrow\) indicate lower values are better.
  When \(k>0\), \(k\) answer generations are sampled from the model---\(\mu, \sigma, \text{and} \gamma\) are the mean, standard deviation and maximum of the \(k\) scores, respectively, averaged over the dataset.
  Otherwise, 1 answer generation is sampled and the mean \(\mu\) is shown.
}
\label{tab:full-consensus-human}}
\vspace*{0.5em}
\end{table*}%

\begin{table*}[!p]
\centering

\def\s{\hspace*{3ex}}
\def\j{0.8ex}
\def\mc#1#2{\multicolumn{#1}{c}{{\sc #2}}}
\def\up{\(\uparrow\)}
\def\down{\(\downarrow\)}
\scalebox{0.58}{
\setlength{\tabcolsep}{3pt}
\begin{tabular}{@{}l*{4}{c}@{\s}*{4}{c}@{\s}c@{\s}*{2}{c}@{\s}*{2}{c}@{}}
  \toprule
  Model                  & \mc{4}{cider\up}                  & \mc{4}{bleu\up}                   & {\sc meteor\up} & \mc{2}{bert}    & \mc{2}{FastText} \\
                         \cmidrule(lr){2-5}                  \cmidrule(lr){6-9}                                 \cmidrule(lr){11-12}\cmidrule(lr){13-14}
                         & n=1    & n=2    & n=3    & n=4    & n=1    & n=2    & n=3    & n=4    &              & \(L_2\)\down& CS\up     & \(L_2\)\down& CS\up      \\
  \midrule
\(\gtA\) \hfill \(\mu\)         & 0.3502 & 0.2479 & 0.2004 & 0.1692 & 0.9948 & 0.4827 & 0.3935 & 0.2940 & 0.9955 & 4.8643 &  0.9215 &  2.2155 &  0.6764 \\
%\hfill \(\sigma\)               & (0.2343) & (0.2041) & (0.1838) & (0.1658) & (0.0498) & (0.4971) & (0.4862) & (0.4512) & (0.0519) & (1.5236) &  (0.0418) &  (0.8601) &  (0.1343) \\
\(\Gamma_H\) \hfill \(\mu\)        & 0.3454 & 0.2734 & 0.2320 & 0.1957 & 1.0000 & 0.7586 & 0.6371 & 0.3816 & 1.0000 & 4.8088 &  0.9220 &  1.9997 &  0.6969 \\
\hfill \(\sigma\)               & (0.2241) & (0.1906) & (0.1682) & (0.1485) & (0.0000) & (1.4082) &  (0.0369) &  (0.6757) &  (0.1070) & (1.4082) &  (0.0369) &  (0.6757) &  (0.1070) \\
\hfill \(\gamma\)               & 0.4212 & 0.3429 & 0.2991 & 0.2583 & 1.0000 & 0.9915 & 0.9767 & 0.7904 & 1.0000 & 4.2891 &  0.9373 &  1.6518 &  0.7614 \\
\cmidrule{2-14}
\gls{CCA}\textsc{-aq-g} (k=1) \hfill \(\mu\)    & 0.0789 & 0.0461 & 0.0313 & 0.0235 & 0.3123 & 0.0752 & 0.0129 & 0.0024 & 0.1864 & 7.1873 &  0.8673 &  3.0908 &  0.4782  \\
\hfill \(\sigma\)               & (0.1634) & (0.1040) & (0.0716) & (0.0544) & (0.3619) & (0.2381) & (0.0962) & (0.0420) & (0.2478) & (1.6917) &  (0.0610) &  (0.9044) &  (0.1639)\\[\j]
% \hfill \(\gamma\)               & 0.0789 & 0.0461 & 0.0313 & 0.0235 & 0.3123 & 0.0752 & 0.0129 & 0.0024 & 0.1864 & 7.1873 &  0.8673 &  3.0908 &  0.4782 \\[\j
%]
\gls{CCA}\textsc{-aq-g} (k=5) \hfill \(\mu\)   & 0.0824 & 0.0489 & 0.0338 & 0.0255 & 0.3244 & 0.1017 & 0.0283 & 0.0048 & 0.2098 & 7.0948 &  0.8690 &  2.9174 &  0.5036 \\
\hfill \(\sigma\)               &(0.1490) & (0.0927) & (0.0645) & (0.0490) & (0.3223) & (0.2113) & (0.0946) & (0.0332) & (0.2355) & (1.3075) &  (0.0451) &  (0.7324) &  (0.1382) \\
\hfill \(\gamma\)               &0.1301 & 0.0805 & 0.0567 & 0.0431 & 0.4967 & 0.2220 & 0.0868 & 0.0182 & 0.3321 & 6.1208 &  0.9015 &  2.4655 &  0.5891 \\[\j]
\gls{CCA}\textsc{-aq-g} (k=10) \hfill \(\mu\)  & 0.0825 & 0.0489 & 0.0342 & 0.0259 & 0.3280 & 0.1041 & 0.0370 & 0.0074 & 0.2255 & 7.0290 &  0.8704 &  2.8408 &  0.5150  \\
\hfill \(\sigma\)               & (0.1397) & (0.0850) & (0.0608) & (0.0462) & (0.3142) & (0.1908) & (0.1062) & (0.0342) & (0.2411) & (1.1715) &  (0.0393) &  (0.6733) &  (0.1279) \\
\hfill \(\gamma\)               & 0.1617 & 0.1025 & 0.0743 & 0.0571 & 0.5902 & 0.3192 & 0.1663 & 0.0477 & 0.4282 & 5.7403 &  0.9118 &  2.2635 &  0.6349 \\[\j]
\gls{CCA}\textsc{-aq-g} (k=15) \hfill \(\mu\)  & 0.0834 & 0.0490 & 0.0345 & 0.0261 & 0.3369 & 0.1009 & 0.0402 & 0.0075 & 0.2431 & 6.9513 &  0.8725 &  2.8181 &  0.5198 \\
\hfill \(\sigma\)               &(0.1295) & (0.0761) & (0.0544) & (0.0413) & (0.3099) & (0.1661) & (0.0963) & (0.0304) & (0.2463) &  (1.1312) &  (0.0377) &  (0.6271) &  (0.1188) \\
\hfill \(\gamma\)               &0.1929 & 0.1240 & 0.0914 & 0.0704 & 0.6572 & 0.3860 & 0.2230 & 0.0647 & 0.5126 & 5.5126 &  0.9172 &  2.1557 &  0.6610 \\
\cmidrule{2-14}
\textsc{hrea-qih-g} (k=1) \hfill \(\mu\) & 0.1109 & 0.0597 & 0.0409 & 0.0308 & 0.4521 & 0.0656 & 0.0243 & 0.0058 & 0.3710 & 6.2743 &  0.8924 &  2.8815 &  0.5334 \\
\hfill \(\sigma\)               & (0.1591) & (0.0909) & (0.0647) & (0.0494) & (0.4281) & (0.2022) & (0.1377) & (0.0672) & (0.4098) & (1.6571) &  (0.0540) &  (0.7971) &  (0.1415) \\[\j]
\textsc{hrea-qih-g} (k=5) \hfill \(\mu\)     & 0.0468 & 0.0251 & 0.0170 & 0.0128 & 0.3457 & 0.0483 & 0.0119 & 0.0022 & 0.2672 & 6.6069 &  0.8809 &  2.8812 &  0.5154 \\
\hfill \(\sigma\)               & (0.0614) & (0.0338) & (0.0233) & (0.0176) & (0.3040) & (0.1102) & (0.0591) & (0.0251) & (0.2738) & (1.3847) &  (0.0474) &  (0.6355) &  (0.1098) \\
\hfill \(\gamma\)               & 0.1518 & 0.0822 & 0.0560 & 0.0422 & 0.5748 & 0.1257 & 0.0351 & 0.0067 & 0.4644 & 5.7237 &  0.9100 &  2.4585 &  0.5980 \\[\j]
\textsc{hrea-qih-g} (k=10) \hfill \(\mu\) & 0.0254 & 0.0136 & 0.0092 & 0.0069 & 0.3134 & 0.0432 & 0.0099 & 0.0017 & 0.2387 &  6.7067 &  0.8778 &  2.8971 &  0.5046 \\
\hfill \(\sigma\)               & (0.0324) & (0.0178) & (0.0122) & (0.0092) & (0.2770) & (0.0961) & (0.0493) & (0.0207) & (0.2444) & (1.2885) &  (0.0448) &  (0.6059) &  (0.1029) \\
\hfill \(\gamma\)               & 0.1569 & 0.0850 & 0.0579 & 0.0436 & 0.5901 & 0.1349 & 0.0365 & 0.0068 & 0.4742 & 5.6594 &  0.9119 &  2.4015 &  0.6059 \\[\j]
\textsc{hrea-qih-g} (k=15) \hfill \(\mu\)& 0.0176 & 0.0094 & 0.0064 & 0.0048 & 0.2962 & 0.0408 & 0.0091 & 0.0016 & 0.2237 & 6.7524 &  0.8765 &  2.9010 &  0.4991 \\
\hfill \(\sigma\)               & (0.0222) & (0.0122) & (0.0084) & (0.0063) & (0.2621) & (0.0898) & (0.0449) & (0.0194) & (0.2275) & (1.2328) &  (0.0433) &  (0.5885) &  (0.0987) \\
\hfill \(\gamma\)               & 0.1590 & 0.0861 & 0.0586 & 0.0442 & 0.5964 & 0.1395 & 0.0372 & 0.0068 & 0.4793 & 5.6348 &  0.9127 &  2.3727 &  0.6096 \\
\cmidrule{2-14}
\textsc{hrea-qih-g*} (k=1) \hfill \(\mu\)      & 0.1580 & 0.0835 & 0.0568 & 0.0428 & 0.5923 & 0.0418 & 0.0221 & 0.0047 & 0.5269 & 5.7023 &  0.9097 &  3.1888 &  0.5196 \\
\hfill \(\sigma\)                &(0.1824) & (0.1048) & (0.0752) & (0.0580) & (0.4656) & (0.1860) & (0.1431) & (0.0631) & (0.4654) & (1.4204) &  (0.0399) &  (0.7521) &  (0.1525) \\[\j]
\textsc{hrea-qih-g*} (k=5)  \hfill \(\mu\)      & 0.1151 & 0.0717 & 0.0523 & 0.0401 & 0.4643 & 0.1551 & 0.0833 & 0.0249 & 0.3553 & 6.0328 &  0.9017 &  2.8978 &  0.5236 \\
\hfill \(\sigma\)                & (0.1033) & (0.0727) & (0.0560) & (0.0433) & (0.2839) & (0.1886) & (0.1418) & (0.0741) & (0.2428) & (1.0778) &  (0.0344) &  (0.5026) &  (0.0983) \\
\hfill \(\gamma\)               & 0.2775 & 0.1804 & 0.1365 & 0.1062 & 0.8505 & 0.4597 & 0.2952 & 0.1038 & 0.7515 & 5.0467 &  0.9253 &  2.1553 &  0.6626 \\[\j]
\textsc{hrea-qih-g*} (k=10) \hfill \(\mu\) & 0.0988 & 0.0624 & 0.0456 & 0.0351 & 0.4225 & 0.1580 & 0.0801 & 0.0264 & 0.3031 &  6.1834 &  0.8978 &  2.8438 &  0.5210 \\
\hfill \(\sigma\)               &(0.0847) & (0.0579) & (0.0436) & (0.0338) & (0.2609) & (0.1621) & (0.1110) & (0.0634) & (0.1986) &  (0.9805) &  (0.0326) &  (0.4866) &  (0.0895) \\
\hfill \(\gamma\)               & 0.3105 & 0.2122 & 0.1653 & 0.1302 & 0.8918 & 0.6207 & 0.4368 & 0.1742 & 0.7991 & 4.8802 &  0.9287 &  2.0014 &  0.6934 \\[\j]
\textsc{hrea-qih-g*} (k=15) \hfill \(\mu\)& 0.0903 & 0.0570 & 0.0416 & 0.0320 & 0.3966 & 0.1529 & 0.0744 & 0.0248 & 0.2740 & 6.2511 &  0.8961 &  2.8371 &  0.5181 \\
\hfill \(\sigma\)                & (0.0765) & (0.0511) & (0.0380) & (0.0295) & (0.2483) & (0.1499) & (0.0966) & (0.0556) & (0.1753) & (0.9376) &  (0.0316) &  (0.4945) &  (0.0876) \\
\hfill \(\gamma\)               & 0.3288 & 0.2303 & 0.1816 & 0.1440 & 0.9118 & 0.6948 & 0.5110 & 0.2133 & 0.8235 & 4.7952 &  0.9304 &  1.9413 &  0.7067 \\
\cmidrule{2-14}
\textsc{mna-qih-g*} (k=1) \hfill \(\mu\)  & 0.1515 & 0.0777 & 0.0523 & 0.0393 & 0.5824 & 0.0224 & 0.0117 & 0.0016 & 0.5239 & 5.7056 &  0.9099 &  3.2581 &  0.5119 \\
\hfill \(\sigma\)                & (0.1756) & (0.0938) & (0.0649) & (0.0495) & (0.4717) & (0.1386) & (0.1050) & (0.0354) & (0.4697) & (1.4043) &  (0.0396) &  (0.7169) &  (0.1509) \\[\j]

\textsc{mna-qih-g*} (k=5) \hfill \(\mu\)    & 0.1118 & 0.0692 & 0.0506 & 0.0387 & 0.4651 & 0.1483 & 0.0828 & 0.0215 & 0.3610 & 6.0294 &  0.9018 &  2.9351 &  0.5177 \\
\hfill \(\sigma\)                & (0.0985) & (0.0692) & (0.0537) & (0.0413) & (0.2868) & (0.1797) & (0.1411) & (0.0669) & (0.2515) & (1.0884) &  (0.0348) &  (0.4845) &  (0.0972) \\
\hfill \(\gamma\)               & 0.2716 & 0.1762 & 0.1336 & 0.1036 & 0.8460 & 0.4560 & 0.2974 & 0.0919 & 0.7494 & 5.0591 &  0.9249 &  2.1713 &  0.6577 \\[\j]

\textsc{mna-qih-g*} (k=10) \hfill \(\mu\)     & 0.0964 & 0.0610 & 0.0448 & 0.0345 & 0.4281 & 0.1573 & 0.0826 & 0.0269 & 0.3124 & 6.1781 &  0.8978 &  2.8812 &  0.5151 \\
\hfill \(\sigma\)                & (0.0791) & (0.0547) & (0.0416) & (0.0322) & (0.2663) & (0.1560) & (0.1083) & (0.0606) & (0.2104) & (0.9934) &  (0.0330) &  (0.4721) &  (0.0879) \\
\hfill \(\gamma\)               & 0.3049 & 0.2102 & 0.1650 & 0.1305 & 0.8865 & 0.6382 & 0.4658 & 0.1881 & 0.7981 & 4.8802 &  0.9284 &  1.9987 &  0.6921 \\[\j]

\textsc{mna-qih-g*} (k=15) \hfill \(\mu\)    & 0.0875 & 0.0554 & 0.0405 & 0.0312 & 0.4007 & 0.1524 & 0.0765 & 0.0257 & 0.2813 & 6.2491 &  0.8960 &  2.8739 &  0.5125 \\
\hfill \(\sigma\)                &(0.0705) & (0.0475) & (0.0356) & (0.0276) & (0.2527) & (0.1424) & (0.0927) & (0.0543) & (0.1858) & (0.9503) &  (0.0321) &  (0.4770) &  (0.0854) \\
\hfill \(\gamma\)               & 0.3221 & 0.2279 & 0.1813 & 0.1442 & 0.9060 & 0.7126 & 0.5459 & 0.2321 & 0.8216 & 4.8014 &  0.9300 &  1.9410 &  0.7049 \\
\cmidrule{2-14}

\textsc{hciae-g-dis} (k=1) \hfill \(\mu\) & 0.1614 & 0.0878 & 0.0605 & 0.0457 & 0.5983 & 0.0730 & 0.0348 & 0.0082 & 0.5138 & 5.7374 &  0.9087 &  3.0389 &  0.5347 \\
\hfill \(\sigma\) & (0.1859) & (0.1125) & (0.0827) & (0.0644) & (0.4497) & (0.2314) & (0.1716) & (0.0812) & (0.4533) & (1.4411) &  (0.0409) &  (0.7878) &  (0.1508) \\
\hfill \(\gamma\) & 0.1614 & 0.0878 & 0.0605 & 0.0457 & 0.5983 & 0.0730 & 0.0348 & 0.0082 & 0.5138 & 5.7374 &  0.9087 &  3.0389 &  0.5347 \\[\j]

\textsc{hciae-g-dis} (k=5) \hfill \(\mu\) & 0.1551 & 0.0845 & 0.0581 & 0.0440 & 0.5791 & 0.0746 & 0.0339 & 0.0083 & 0.4939 & 5.7890 &  0.9073 &  3.0136 &  0.5351 \\
\hfill \(\sigma\) & (0.1643) & (0.0966) & (0.0699) & (0.0542) & (0.4043) & (0.1837) & (0.1252) & (0.0582) & (0.4090) & (1.3565) &  (0.0388) &  (0.6998) &  (0.1349) \\
\hfill \(\gamma\) & 0.2063 & 0.1173 & 0.0825 & 0.0629 & 0.7082 & 0.1626 & 0.0863 & 0.0239 & 0.6111 & 5.4373 &  0.9168 &  2.7232 &  0.5840 \\[\j]

\textsc{hciae-g-dis} (k=10) \hfill \(\mu\) & 0.1551 & 0.0846 & 0.0582 & 0.0440 & 0.5789 & 0.0749 & 0.0338 & 0.0084 & 0.4935 & 5.7920 &  0.9073 &  3.0149 &  0.5349 \\
\hfill \(\sigma\) & (0.1617) & (0.0951) & (0.0686) & (0.0532) & (0.3979) & (0.1773) & (0.1197) & (0.0560) & (0.4032) & (1.3447) &  (0.0385) &  (0.6881) &  (0.1330) \\
\hfill \(\gamma\) & 0.2243 & 0.1301 & 0.0924 & 0.0706 & 0.7488 & 0.2083 & 0.1145 & 0.0347 & 0.6480 & 5.3431 &  0.9192 &  2.6187 &  0.6010 \\[\j]

\textsc{hciae-g-dis} (k=15) \hfill \(\mu\) & 0.1551 & 0.0846 & 0.0582 & 0.0440 & 0.5791 & 0.0750 & 0.0340 & 0.0084 & 0.4931 & 5.7927 &  0.9072 &  3.0142 &  0.5351 \\
\hfill \(\sigma\) & (0.1607) & (0.0944) & (0.0681) & (0.0528) & (0.3959) & (0.1752) & (0.1172) & (0.0543) & (0.4014) & (1.3402) &  (0.0384) &  (0.6837) &  (0.1323) \\
\hfill \(\gamma\) & 0.2342 & 0.1377 & 0.0985 & 0.0755 & 0.7679 & 0.2365 & 0.1371 & 0.0428 & 0.6669 & 5.2942 &  0.9204 &  2.5596 &  0.6106 \\
\cmidrule{2-14}

\textsc{rva} (k=1) \hfill \(\mu\)  & 0.1209 & 0.0650 & 0.0445 & 0.0336 & 0.4850 & 0.0673 & 0.0256 & 0.0066 & 0.4033 & 6.1629 &  0.8956 &  2.9040 &  0.5353 \\
\hfill \(\sigma\)     &(0.1644) & (0.0941) & (0.0678) & (0.0521) & (0.4378) & (0.2101) & (0.1431) & (0.0722) & (0.4238) & (1.6415) &  (0.0527) &  (0.7994) &  (0.1437) \\[\j]
%\hfill \(\gamma\)   & 0.1209 & 0.0650 & 0.0445 & 0.0336 & 0.4850 & 0.0673 & 0.0256 & 0.0066 & 0.4033 & 6.1629 &  0.8956 &  2.9040 &  0.5353 \\
%
\textsc{rva}  (k=5) \hfill \(\mu\)  & 0.1207 & 0.0650 & 0.0445 & 0.0336 & 0.4817 & 0.0682 & 0.0269 & 0.0069 & 0.4028 & 6.1633 &  0.8955 &  2.9017 &  0.5358 \\
\hfill \(\sigma\)     &(0.1243) & (0.0678) & (0.0475) & (0.0362) & (0.3351) & (0.1266) & (0.0799) & (0.0375) & (0.3314) & (1.3054) &  (0.0404) &  (0.5718) &  (0.1062) \\
\hfill \(\gamma\)   & 0.2165 & 0.1234 & 0.0868 & 0.0661 & 0.7490 & 0.2239 & 0.1055 & 0.0304 & 0.6422 & 5.3442 &  0.9195 &  2.3333 &  0.6348 \\[\j]
\textsc{rva}  (k=10) \hfill \(\mu\)   & 0.1210 & 0.0651 & 0.0446 & 0.0337 & 0.4830 & 0.0682 & 0.0268 & 0.0068 & 0.4032 & 6.1638 &  0.8955 &  2.9009 &  0.5358 \\
\hfill \(\sigma\)     & (0.1185) & (0.0639) & (0.0444) & (0.0338) & (0.3200) & (0.1120) & (0.0664) & (0.0298) & (0.3172) & (1.2562) &  (0.0387) &  (0.5358) &  (0.1000) \\
\hfill \(\gamma\)   & 0.2525 & 0.1496 & 0.1073 & 0.0821 & 0.8163 & 0.3327 & 0.1727 & 0.0528 & 0.7066 & 5.1729 &  0.9236 &  2.1777 &  0.6638 \\[\j]
\textsc{rva}  (k=15) \hfill \(\mu\)   & 0.1211 & 0.0652 & 0.0447 & 0.0337 & 0.4826 & 0.0685 & 0.0271 & 0.0070 & 0.4028 & 6.1603 &  0.8956 &  2.9008 &  0.5360 \\
\hfill \(\sigma\)     &(0.1164) & (0.0626) & (0.0435) & (0.0330) & (0.3142) & (0.1079) & (0.0632) & (0.0275) & (0.3121) & (1.2403) &  (0.0380) &  (0.5230) &  (0.0982) \\
\hfill \(\gamma\)   & 0.2709 & 0.1646 & 0.1196 & 0.0919 & 0.8457 & 0.4002 & 0.2239 & 0.0736 & 0.7376 & 5.0869 &  0.9256 &  2.1026 &  0.6778 \\
\bottomrule
\end{tabular}}
\vspace*{-1ex}
{\caption{
  Overlap and embedding distance metrics computed for \(k\) generations against the automatic reference sets \(\Sigma\) on the \emph{entire} validation set.
  For \textsc{hrea-qih-g}, on average \(\sim\)6 answers are the empty string, which are excluded in computing statistics.
  Metrics marked~\(\uparrow\) indicate higher values are better, and those marked~\(\downarrow\) indicate lower values are better.
  When \(k > 0\), \(k\) answer generations are sampled from the model---\(\mu, \sigma, \text{and} \gamma\) are the mean, standard deviation and maximum of the \(k\) scores, respectively, averaged over the dataset.
  Otherwise, 1 answer generation is sampled and the mean \(\mu\) is shown.
}\label{tab:full-consensus-auto}}
\vspace*{0.5em}
\end{table*}

In~\cref{tab:consensus} in the main paper, we show the performance of models using the revised evaluation scheme.
Specifically, we measure the similarity of generated answers against either human-annotated reference sets \(H\) (for \(\mathcal{H}_v\)) or automatic reference sets \(\Sigma\) (for the entire validation set).
We measure similarity using the overlap and embedding distance-based metrics described in the main paper.

In~\cref{tab:full-consensus-human} and~\cref{tab:full-consensus-auto}, we extend this analysis for \(H\) and \(\Sigma\) respectively, showing the empirical results for an increasing number of \(k=\{1,5,10,15\}\) generations sampled from each model.
We show the mean \(\mu\), standard deviation \(\sigma\), and maximum \(\gamma\) across the \(k\) scores.
Note, this is visually presented in~\cref{fig:ksample} in the main paper.

We also include two baselines intended as upper bounds:
\begin{inparaenum}[i)]
\item \(\gtA\), takes the ground-truth answer to be the generated answer, and
\item \(\Gamma_H\), cycles through \(H\), treating each answer as the generated answer. Since each answer in the set could be a plausible one (as marked by humans), we take the best-case score (minimum score for all except embedding-based \(L_2\) distance for which we take the maximum), and then average over the dataset.
\end{inparaenum}

\section{Automatic reference set construction}
\label{sec:app-clustering}

In our work, we explore different semi-supervised methods for automatically extracting relevant answers from given candidate sets, based on the questions and images.
These extracted reference sets are then used to evaluate answer generations for the entire \textit{VisDial} dataset.

As described in the main paper, we apply \gls{CCA}\textsc{-aq*} to solely those \((Q,A)\) pairs in \(\mathcal{H}_t\) for which humans scored \(\rho(A)>0\), and then compute the correlations \(\mathcal{C}\) between~\(\gtA\) and each \(A \in \tilde{\bm{A}}\), where \(\tilde{\bm{A}} = \bm{A} \setminus \gtA\).
We then explore the following clustering heuristics on \(\mathcal{C}\):
\begin{compactdesc}
\item[Simple:]  \(\Sigma = \{A : \phi(\gtA, A) \in [\mathcal{C}_{\text{max}} - \sigma, \mathcal{C}_{\text{max}}]\} \cup \{\gtA\}\), where~\(\mathcal{C}_{\text{max}} = \mathrm{max}(\mathcal{C}), \sigma = \mathrm{stdev}(\mathcal{C})\).
\item[Meanshift:] choosing the best-ranked cluster~\(M'\) after running meanshift~\citep{comaniciu2002mean} on~\(\mathcal{C}\) to derive~\(M = M' \cup \{\gtA\}\).
\item[Agglomerative:] choosing the best-ranked cluster~\(G'\) after running agglomerative clustering on~\(\mathcal{C}\), with number of clusters set to 5, to derive~\(G = G' \cup \{\gtA\}\).
\end{compactdesc}
Note, each unions the resulting set with \(\gtA\).

\begin{figure}
\includegraphics[width=0.5\textwidth]{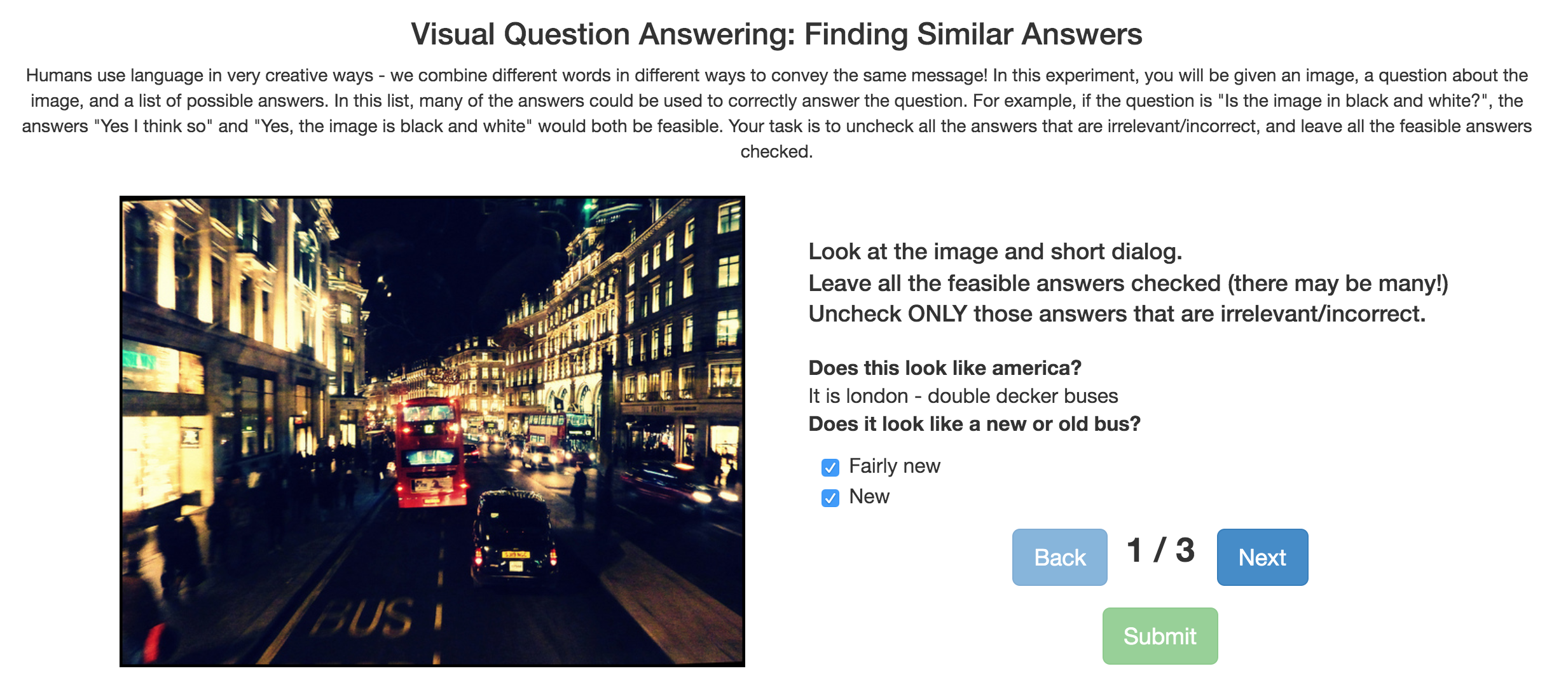}\\
\includegraphics[width=0.5\textwidth]{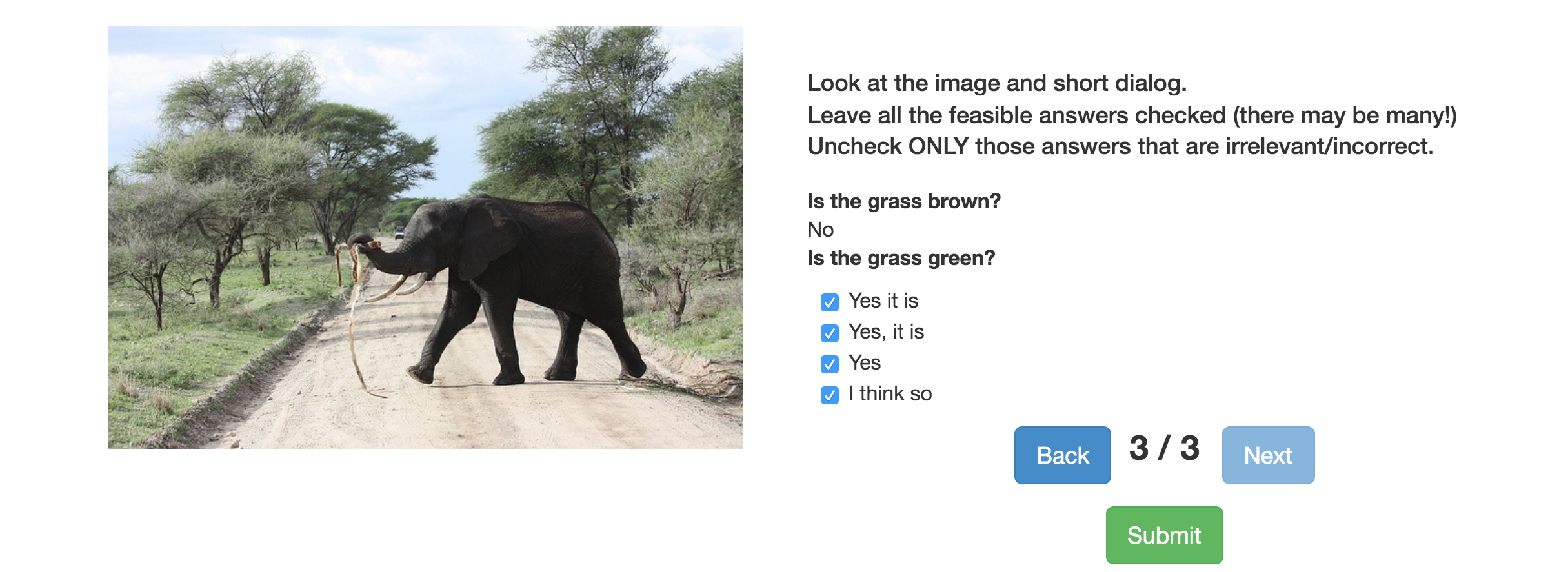}\\
\includegraphics[width=0.5\textwidth]{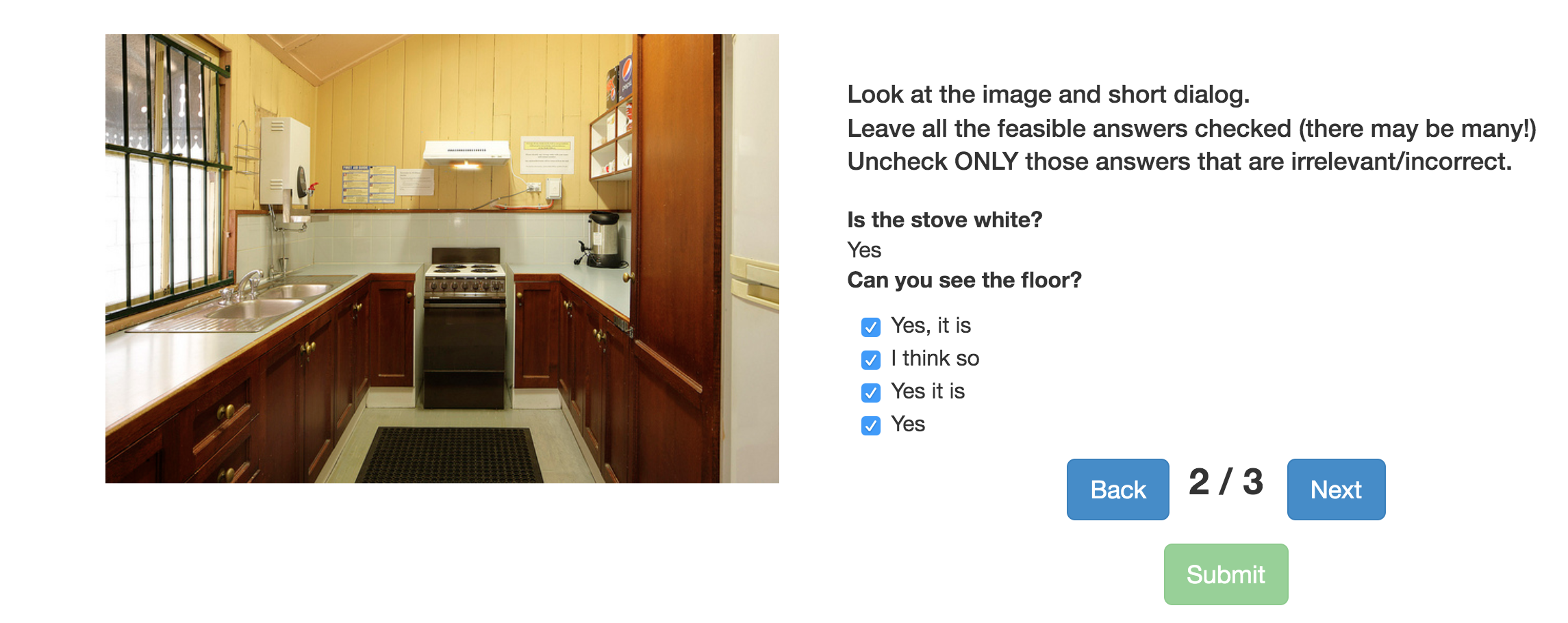}\\
\includegraphics[width=0.5\textwidth]{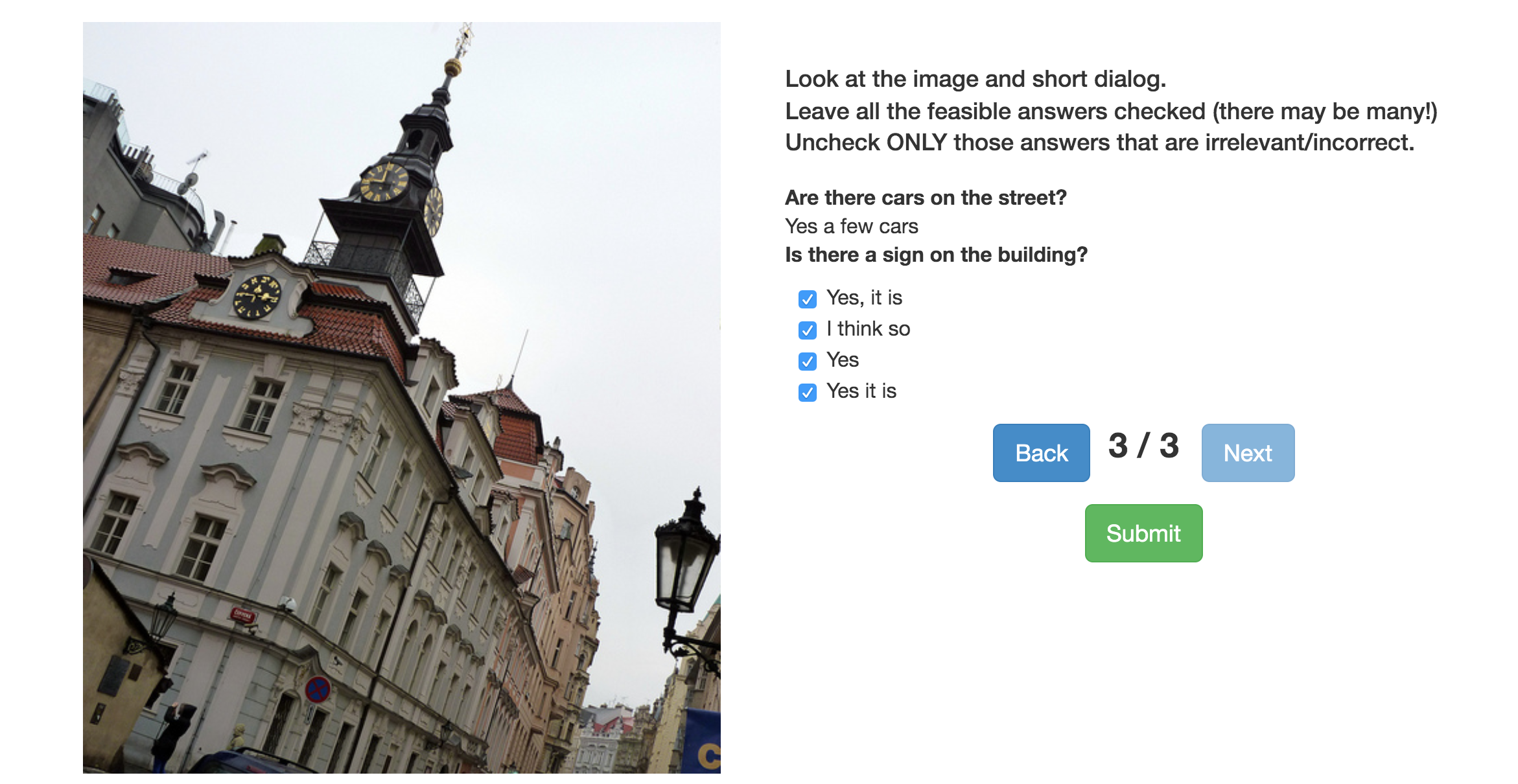}
\caption{
  The \gls{AMT} interface we use to verify answer reference sets, sampled from \(H\) and \(\Sigma\).
  Turkers are asked to \emph{de-select} the irrelevant or incorrect answers given an image and question.
}
\label{fig:amt-interface}
\end{figure}

\begin{table}%[!ht]
  \centering
  \def\s{\hspace*{4ex}}
  \newcommand{\mr}[2]{\multirow{#1}{*}{{#2}}}
  \newcommand{\mc}[2]{\multicolumn{#1}{c}{{#2}}}
  \scalebox{0.6}{
  \setlength{\tabcolsep}{3pt}
  \begin{tabular}{@{}cr>{$}c<{$}@{\s}*{6}{c}@{}}
    \toprule
    \mc{2}{Train}               & \mr{2}{Ref} & \mr{2}{MR} & \mr{2}{R@1} & \mr{2}{R@5} & \mr{2}{R@10} & \mr{2}{MRR} & \mr{2}{NDCG} \\
    \cmidrule{1-2}
    \text{Set}     & \#QA pairs &           &            &             &             &              &             &  \\
    \midrule
    \mr{5}{\(\mathcal{H}_t\)}
      & 15,317     & H         & 26.49      & 6.05        & 21.50       & 35.53        & 0.1550      & 0.3647 \\
      & 17,055     & \Sigma    & 20.36      & 8.35        & 32.88       & 48.78        & 0.2066      & 0.3715 \\
      & 26,318     & M         & 21.53      & 6.83        & 29.80       & 45.63        & 0.1862      & 0.3503\\
      & 16,923     & G \text{ (n=5)} & 20.66 & 8.08       & 30.35       & 46.33        & 0.1981      & 0.3657\\
      & 1996       & \{\gtA\}  & 23.71      & 13.13       & 34.05       & 46.90        & 0.2428      & 0.2734 \\
    \cmidrule(l){2-9}
    \mr{4}{all train}
      & 1,232,870  & \{\gtA\}  & 17.07      & 16.18       & 40.18       & 55.35        & 0.2845      & 0.3493 \\
      & 10,419,489 & \Sigma    & 17.20      & 10.73       & 34.20       & 51.80        & 0.2312      & 0.4023 \\
      & 17,600,151 & M     & 20.67          & 9.38        & 24.45       & 39.93        & 0.1905      & 0.3339 \\
      & 10,614,163 & G \text{ (n=5)} & 17.79  & 9.78      & 31.40         & 48.93         & 0.2171    & 0.3918\\
    \mr{4}{all trainval}
      & 1,253,510 & \{\gtA\}    & 17.10     & 16.10       & 40.05       & 55.07        & 0.2833      & 0.3486\\
      & 10,599,533 & \Sigma     & 17.31     & 10.20       & 33.30       & 51.45        & 0.2242      & 0.4050 \\
      & 17,931,897 & M          & 20.47     & 9.38        & 24.83       & 40.55        & 0.1917      & 0.3380 \\
      & 10,798,877 & G \text{ (n=5)} & 17.60      & 9.85  & 31.67       & 49.20        & 0.2184      & 0.3927\\
    \bottomrule
  \end{tabular}}
  \vspace*{-1ex}
  {\caption
  {%
    Extended evaluation of the utility of automatic reference set construction methods (\(M, \Sigma, G\)) on the standard \gls{VD} evaluation.
    Models were trained using \gls{CCA} on the indicated subsets (\(\mathcal{H}_t\), all train, or all trainval) of \textit{VisDial v1.0}, using answers drawn from different sets (`Ref'), and tested on the evaluation \href{https://visualdialog.org/challenge/2018}{test server}.
  }
  \label{tab:app-ndcg-cluster}}
\end{table}

\begin{table}%[!ht]
  \centering
  \def\s{\hspace*{4ex}}
  \newcommand{\mr}[2]{\multirow{#1}{*}{{#2}}}
  \newcommand{\mc}[2]{\multicolumn{#1}{c}{{#2}}}
  \scalebox{0.6}{
  \setlength{\tabcolsep}{3pt}
  \begin{tabular}{@{}cr>{$}c<{$}@{\s}*{6}{c}@{}}
    \toprule
    \mc{2}{Train}               & \mr{2}{Ref} & \mr{2}{MR} & \mr{2}{R@1} & \mr{2}{R@5} & \mr{2}{R@10} & \mr{2}{MRR} & \mr{2}{NDCG} \\
    \cmidrule{1-2}
    \text{Set}     & \#QA pairs &           &            &             &             &              &             &  \\
    \midrule
    \mr{5}{\(\mathcal{H}_t\)}
      & 15,317     & H         & 25.34      & 6.20      & 22.58         & 37.84      & 0.1598       & 0.3755       \\
      & 17,055     & \Sigma    & 20.94      & 8.54      & 32.16         & 47.69      & 0.2049       & 0.3884       \\
      & 26,318     & M         & 21.84      & 7.16      & 28.78         & 44.95      & 0.1858       & 0.3669      \\
      & 16,923     & G \text{ (n=5)}& 21.47 & 7.93      & 30.29         & 45.89      & 0.1942       & 0.3779      \\
      & 1996       & \{\gtA\}  & 23.80      & 13.50     & 34.06         & 46.64      & 0.2442       & 0.2816       \\
    \cmidrule(l){2-9}
    \mr{4}{all train}
      & 1,232,870  & \{\gtA\}  & 17.04      & 16.00     & 41.21         & 55.16      & 0.2860       & 0.3547       \\
      & 10,419,489 & \Sigma    & 17.39      & 10.27     & 34.01         & 51.54      & 0.2264       & 0.4099    \\
      & 17,600,151 & M         & 20.96      & 9.11      & 22.92         & 39.30      & 0.1850       & 0.3354 \\
      & 10,614,163 & G \text{ (n=5)} & 18.03 & 9.68   & 31.16           & 48.85      & 0.2136       & 0.4005 \\
    %\mr{4}{all trainval} cannot test on val if trained on val!
    \bottomrule
  \end{tabular}}
  \vspace*{-1ex}
  {\caption
  {%
    Extended evaluation of the utility of automatic reference set construction methods (\(M, \Sigma, G\)) on the standard \gls{VD} evaluation.
    Models were trained using \gls{CCA} on the indicated subsets (\(\mathcal{H}_t\), all train, or all trainval) of \textit{VisDial v1.0}, using answers drawn from different sets (`Ref'), and tested on the validation set.
  }
  \label{tab:app-ndcg-cluster-val}}
\end{table}

\subsection{Verifying automatic reference sets}

In the main paper, we describe a three-step procedure which we use to verify the quality of the automatic reference sets.

\paragraph{Computing intersection with \(H\)}

\cref{tab:app-clustering} extends~\cref{tab:clusters} in the main paper and shows the \gls{IOU}, precision, recall, and set size for the reference sets extracted by the different methods \(C=\{\Sigma, M, G\}\) against \(H\)---denoted \((\gtA, \tilde{\bm{A}})\).
Within the reference sets \(C\), we also compute the average correlation, the standard deviation of the correlations, and the likelihood of \(C\) containing \(\gtA\).

In~\cref{tab:app-clustering} we additionally show the results of the above, but using
\begin{inparaenum}[i)]
\item \gls{CCA}\textsc{-aq}, learned on \emph{all} train \((Q,A)\) pairs, and
\item correlations \(\mathcal{C}\) computed between \(\gtA\) or the question \(Q\), and the \emph{full} candidate set---\((\gtA, \bm{A})\) and \((Q, \bm{A})\), respectively.
\end{inparaenum}
In the case of (\(Q, \bm{A}\)), we can construct \(C\) in two ways: either by selecting all those answers in \(\bm{A}\) with the same cluster label as \(\gtA\), or those with the same label as the answer with the maximum correlation to \(Q\). We denote this \((Q,\bm{A})_{\text{gt}}\) and \((Q,\bm{A})_{\text{max}}\), respectively.
This does not apply to \((\gtA,\bm{A})\) since \(\gtA\) and the answer with the maximum correlation will always be the same, nor for \((\gtA,\tilde{\bm{A}})\), since \(\gtA\), by definition, is excluded from \(\tilde{\bm{A}}\), and simply unioned with the resulting cluster afterwards.

\begin{figure*}
  \centering\tiny
  \def\sb{1.08}
  \renewcommand*{\arraystretch}{1.1}
  \begin{tabular}{@{}l@{\hspace*{1ex}}l@{}l@{}l@{}}
    \includegraphics[trim={0 0 0 0.73cm},clip,width=0.23\linewidth]{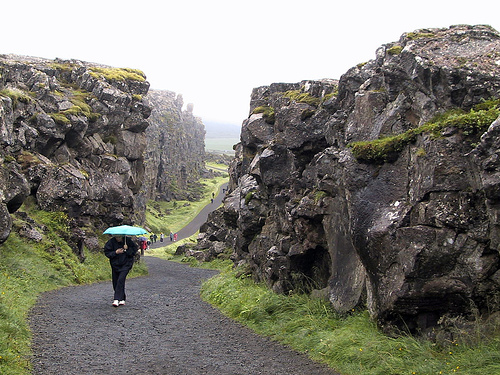}
    & \includegraphics[width=0.23\linewidth]{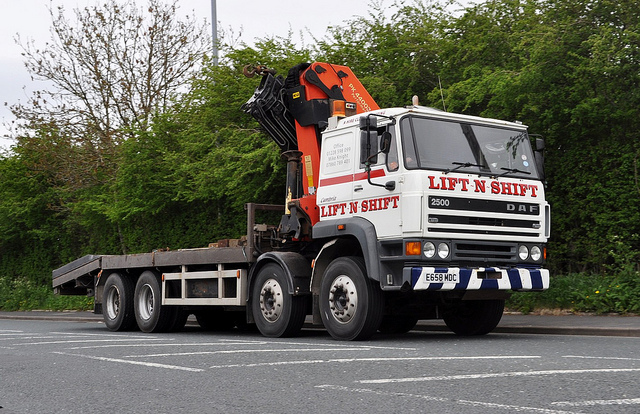}
    & \includegraphics[width=0.23\linewidth]{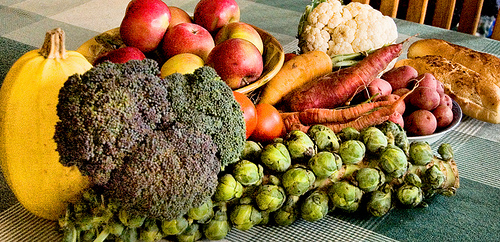}
    & \includegraphics[trim={0 3.5cm 0 8cm},clip,width=0.23\linewidth]{VisualDialog_train_000000088029}
    \\[1ex]
    \scalebox{\sb}{%
    \begin{tabular}[t]{@{}r@{\,}p{0.21\linewidth}@{}}
      Q: & Are there any other people? \\
         & \rank[purple]{GT} \textcolor{purple}{Way in the background} \\
      \parbox[t]{2mm}{\multirow{4}{*}}
         & \rank{} There are few people way off in background \\
         & \rank{} I see a few in the background \\
         & \rank{} There are a few in the background
    \end{tabular}}
    &
    \scalebox{\sb}{%
    \begin{tabular}[t]{@{}r@{\,}p{0.21\linewidth}@{}}
      Q: & Is the driver of the truck nearby? \\
         & \rank[purple]{GT} \textcolor{purple}{I can't see anyone in the picture} \\
      \parbox[t]{2mm}{\multirow{4}{*}}
         & \rank{} No people \\
         & \rank{} Can't see anyone else
    \end{tabular}}
    &
    \scalebox{\sb}{%
    \begin{tabular}[t]{@{}r@{\,}p{0.21\linewidth}@{}}
      Q: & Is the broccoli raw or cooked? \\
         & \rank[purple]{GT} \textcolor{purple}{It's raw.} \\
      \parbox[t]{2mm}{\multirow{4}{*}}
         & \rank{} Raw
    \end{tabular}}
    &
    \scalebox{\sb}{%
    \begin{tabular}[t]{@{}r@{\,}p{0.21\linewidth}@{}}
      Q: & Is the mountain large or small? \\
         & \rank[purple]{GT} \textcolor{purple}{It's large} \\
      \parbox[t]{2mm}{\multirow{4}{*}}
         & \rank{} Fairly large \\
         & \rank{} It's medium size\\
         & \rank{} Large\\
         & \rank{} Pretty large\\
         & \rank{} Medium size I would say not small not large\\
         & \rank{} I would say large
    \end{tabular}}
  \end{tabular}
  \vspace*{-1.5em}
  \caption
  {%
    Qualitative examples of the relevant answers our semi-supervised approach (\(\Sigma\)) extracts from given candidate answer sets.
    Note, we show \emph{all} answers which our method extracts from the sets.
  }
  \label{fig:app-cluster-quality}
  \vspace{0.5ex}
\end{figure*}

% \vspace*{-0.5em}
We cross-validate and select the method (in our case, \(\Sigma\)) which gives us the best precision, \(|H \cap C|/|C|\), and a small cluster size, \(|C|\).
Using this method, we show some qualitative examples of the automatic reference sets \(\Sigma\) in~\cref{fig:app-cluster-quality}.
The majority of the answers are relevant both to the image and the question.

\begin{table*}%[ht!]
\centering
\def\s{\hspace*{7pt}}
 \newcommand{\rt}[2]{\parbox[t]{2mm}{\multirow{#1}{*}{\rotatebox[origin=c]{90}{#2}}}}
\scalebox{0.78}{
\begin{tabular}{@{}c@{\s}c@{\s}l@{\s}c@{\s}c@{\s}c@{\s}c@{\s}c@{\s}c@{\s}c@{}}
\toprule
                                C & & & \(\frac{|H \cap C|}{|H \cup C|}\)        &    \(\frac{|H \cap C|}{|C|}\)   &  \(\frac{|H \cap C|}{|H|}\) & \(|C|\) & \(\gtA \in C\) & corr (\(C\)) & std (corr(\(C\)))     \\
\midrule
\(H\) &  & --------
& 100.00 (0.00) & 100.00 (0.00) & 100.00 (0.00) & 12.77 (7.24) & 100.00 (0.00) & 0.2393 (0.1942) & 0.1670 (0.0917)\\
\midrule
\rt{8}{\(M\)}
& \rt{4}{\acrshort{CCA}\textsc{-aq}} & \((Q, \bm{A})_{\text{gt}}\)
& 17.59 (13.04) & 31.01 (27.31) & 57.67 (32.17) & 39.85 (31.51) & 100.00 (0.00) & 0.2998 (0.2610) & 0.0856 (0.0374) \\
%& & & (bw-bw) (Q-100A)
%& 16.44 (11.92) & 26.82 (25.68) & 70.44 (33.18) & 57.81 (39.71) & 100.00 (0.00) & 0.2707 (0.2824) & 0.1029 (0.0301) \\
& & \((Q, \bm{A})_{\text{max}}\)
& 15.06 (14.31) & 37.02 (34.72) & 38.43 (35.62) & 21.53 (26.14) & 54.89 (49.77) & 0.4794 (0.2379) & 0.0852 (0.0424) \\
& & \((\gtA,\bm{A})\)
& 13.14 (12.74) & 89.10 (26.06) & 18.57 (23.21) & 4.37 (10.36) & 100.00 (0.00) & 0.9466 (0.1016) & 0.1586 (0.0844) \\
& & \((\gtA,\tilde{\bm{A}})\)
& 19.05 (12.65) & 42.68 (30.16) & 50.47 (35.70) & 27.96 (32.01) & 100.00 (0.00) & 0.5069 (0.2734) & 0,2068 (0.0939) \\
%& & & (k=50) (GTA-99A)
%& 20.95 (13.19) & 33.95 (24.79) & 61.63 (33.19) & 34.60 (29.59) & 100.00 (0.00) & 0.5587 (0.2539) & 0.1814 (0.0698) \\
%& & & (k=10) (GTA-99A)
%& 19.27 (10.86) & 22.59 (14.21) & 72.14 (27.21) & 42.52 (17.47) & 100.00 (0.00) & 0.6515 (0.1794) & 0.1664 (0.0578) \\
\cmidrule{2-10}
& \rt{4}{\gls{CCA}\textsc{-aq*}}
%& \rt{4}{\(\mu\)(all)}
%& \rt{2}{GT} & CCA-QA (Q-100A)
%& 20.05 (15.86) & 26.53 (23.83) & 65.84 (27.87) & 45.93 (28.88) & 100.00 (0.00) & 0.2001 (0.2122) & 0.0782 (0.0387) \\
%& & & & CCA-QA (bw-bw) (Q-100A)
%& 17.52 (14.22) & 22.47 (22.24) & 79.07 (28.63) & 66.93 (37.63) & 100.00 (0.00) & 0.1766 (0.2240) & 0.0904 (0.0263) \\
%& & & \rt{3}{top} & CCA-QA (Q-100A)
%& 21.10 (17.54) & 33.64 (29.69) & 55.17 (34.64) & 30.72 (27.92) & 59.64 (49.07) & 0.3176 (0.2136) & 0.0770 (0.0390) \\
%& & & & CCA-QA (GTA-100A)
%&  14.12 (13.58) & 93.39 (20.29) & 17.64 (20.95) & 3.29 (8.17) & 100.00 (0.00) & 0.9542 (0.1111) & 0.1297 (0.0621) \\
%& & & & CCA-QA (GTA-99A)
%& 25.99 (18.18) & 54.68 (29.66) & 44.88 (31.06) & 15.13 (19.93) & 100.00 (0.00) & 0.5715 (0.2243) & 0.2111 (0.0953)\\
%& & & & CCA-QA (bw-bw) (GTA-99A)
%& 25.87 (17.89) & 50.64 (32.11) & 55.75 (33.46) & 29.21 (37.29) & 100.00 (0.00) & 0.5112 (0.2862) & 0.1755 (0.0665) \\
%& & & & CCA-QA (k=50) (GTA-99A)
%& 25.80 (17.48) & 38.42 (26.41) & 62.63 (31.69) & 29.66 (25.36) & 100.00 (0.00) & 0.5698 (0.2345) & 0.1835 (0.0639) \\
& \((Q, \bm{A})_{\text{gt}}\)
& 18.66 (14.83) & 25.33 (23.22) & 62.46 (27.37) & 45.89 (28.70) & 100.00 (0.00) & 0.1843 (0.2146) & 0.0735 (0.0360) \\
& & \((Q, \bm{A})_{\text{max}}\)
& 19.89 (16.99) & 34.58 (30.69) & 49.39 (34.08) & 26.74 (25.82) & 53.49 (49.89) & 0.3270 (0.2186) & 0.0710 (0.0367) \\
& & \((\gtA,\bm{A})\)
& 13.15 (12.84) & 93.96 (20.12) & 16.64 (20.51) & 3.12 (8.11) & 100.00(0.00) & 0.9660 (0.0959) & 0.1207 (0.0686) \\
& & \((\gtA,\tilde{\bm{A}})\)
& 25.01 (18.40) & 59.19 (31.55) & 39.35 (28.86) & 12.00 (16.07) & 100.00 (0.00) & 0.5841 (0.2167) & 0.2231 (0.0990) \\
\midrule
\rt{8}{\(\Sigma\)}
& \rt{4}{\acrshort{CCA}\textsc{-aq}}
& \((Q, \bm{A})_{\text{gt}}\)
& 19.92 (12.94) & 32.64 (26.49) & 64.25 (31.71) & 39.45 (29.86) & 100.00 (0.00) & 0.3334 (0.2310) & 0.1162 (0.0637) \\
& & \((Q, \bm{A})_{\text{max}}\)
& 14.42 (13.89) & 40.72 (35.20) & 27.33 (28.47) & 11.32 (13.38) & 47.14 (49.93) & 0.5193 (0.2034) & 0.0661 (0.0295) \\
& & \((\gtA,\bm{A})\)
& 12.67 (11.97) & 89.83 (25.66) & 18.17 (22.76) & 4.01 (9.08) & 100.00 (0.00) & 0.9658 (0.0753) & 0.0960 (0.0355) \\
& & \((\gtA,\tilde{\bm{A}})\)
& 21.16 (13.60) & 49.21 (28.34) & 38.60 (28.03) & 12.30 (12.17) & 100.00 (0.00) & 0.5859 (0.1898) & 0.2092 (0.0949) \\
\cmidrule{2-10}
& \rt{4}{\acrshort{CCA}\textsc{-aq*}}
%& \rt{3}{\(\mu\)(all)}
%& GT & (Q-100A)
%& 23.69 (16.59) & 29.26 (23.14) & 77.83 (23.95) & 47.46 (28.78) & 100.00 (0.00) & 0.2345 (0.1832) & 0.0991 (0.0632) \\
%& & & \rt{2}{top} & CCA-QA (Q-100A)
%& 21.36 (18.09) & 41.39 (32.84) & 35.66 (27.57) & 12.70 (9.77) & 42.05 (49.38) & 0.3743 (0.1936) & 0.0524 (0.0267) \\
%& & & & CCA-QA (GTA-99A)
%& 25.90 (17.74) & 61.94 (30.33) & 35.28 (24.20) & 7.60 (6.91) & 100.00 (0.00) & 0.6370 (0.1781) & 0.2095 (0.0967) \\
& \((Q, \bm{A})_{\text{gt}}\)
& 21.94 (15.26) & 27.31 (22.21) & 77.53 (24.32) & 50.37 (29.80) & 100.00 (0.00) & 0.2194 (0.1879) & 0.1002 (0.0647) \\
& & \((Q, \bm{A})_{\text{max}}\)
& 19.64 (17.18) & 40.95 (33.44) & 32.89 (26.85) & 11.93 (9.73) & 37.02 (48.30) & 0.3739 (0.2000) & 0.0505 (0.0264) \\
& & \((\gtA,\bm{A})\)
& 13.15 (12.89) & 93.32 (19.91) & 15.49 (17.80) & 2.10 (3.72) & 100.00 (0.00) & 0.9764 (0.0596) & 0.0897 (0.0369) \\
& & \((\gtA,\tilde{\bm{A}})\)
& 24.13 (16.73) & 62.48 (31.24) & 32.91 (23.52) & 7.17 (6.94) & 100.00 (0.00) & 0.6206 (0.1773) & 0.2269 (0.0975) \\
% removed (-sigma, +sigma)
\midrule
\rt{12}{\(G\)}
& \rt{7}{\acrshort{CCA}\textsc{-aq}}
& \((Q, \bm{A})_{\text{gt}}\) \hfill n=3
& 19.56 (13.04) & 27.14 (21.46) & 58.03 (26.41) & 33.78 (18.68) & 100.00 (0.00) & 0.2888 (0.2430) & 0.0851 (0.0398) \\
& & \((Q, \bm{A})_{\text{gt}}\) \hfill n=4
& 19.15 (12.81) & 31.61 (24.59) & 46.77 (25.22) & 23.96 (14.68) & 100.00 (0.00) & 0.3193 (0.2488) & 0.0679 (0.0355) \\
& & \((Q, \bm{A})_{\text{gt}}\) \hfill n=5
& 10.49 (6.31) & 17.08 (13.61) & 28.42 (17.82) & 21.25 (8.64) & 100.00 (0.00) & 0.5656 (0.1375) & 0.0287 (0.0100)\\
& & \((Q, \bm{A})_{\text{max}}\) \hfill n=4
& 17.65 (14.16) & 36.87 (29.85) & 35.27 (27.86) & 14.75 (11.63) & 57.17 (49.50) & 0.4716 (0.2189) & 0.0758 (0.0380) \\
& & \((Q, \bm{A})_{\text{max}}\) \hfill n=5
& 9.47 (8.99) & 20.73 (24.04) & 19.14 (17.22) & 14.94 (8.64) & 20.78 (40.59) & 0.7673 (0.0605) & 0.0336 (0.0121)\\
& & \((\gtA,\bm{A})\) \hfill n=5
& 15.53 (13.11) & 81.47 (29.65) & 21.82 (22.55) & 4.70 (7.58) & 100.00 (0.00) & 0.8879 (0.1561) & 0.1513 (0.0692)\\
& & \((\gtA,\tilde{\bm{A}})\) \hfill n=5
& 22.13 (13.18) & 47.40 (27.16) & 38.93 (25.76) & 11.58 (9.13) & 100.00 (0.00) &  0.5633 (0.1943) & 0.1987 (0.0834)\\
\cmidrule{2-10}
& \rt{5}{\gls{CCA}\textsc{-aq*}}
%& \rt{6}{\(\mu\)(all)}
%& GT & n=4 (Q-100A)
%& 21.57 (16.04) & 30.60 (24.74) & 50.43 (24.37) & 25.27 (12.23) & 100.00 (0.00) & 0.2247 (0.2060) & 0.0525 (0.0285) \\
%& & & \rt{5}{top} & n=4 (Q-100A)
%& 24.02 (18.05) & 38.50 (29.29) & 44.65 (27.66) & 16.87 (9.43) & 50.92 (50.00) & 0.3481 (0.2019) & 0.0568 (0.0289) \\
%& & & & n=4 (Q-99A)
%& 26.56 (17.57) & 40.99 (27.58) & 50.57 (25.51) & 17.68 (9.38) & 100.00 (0.00) & 0.3333 (0.1963) & 0.0775 (0.0494) \\
%& & & & n=3 (GTA-99A)
%& 28.00 (17.96) & 41.88 (26.46) & 57.50 (29.03) & 20.64 (13.35) & 100.00 (0.00) & 0.4389 (0.2027) & 0.1932 (0.0530) \\
%& & & & n=4 CCA-QA (GTA-99A)
%& 28.85 (19.02) & 52.31 (29.42) & 46.52 (28.00) & 12.53 (9.09) & 100.00 (0.00) & 0.5368 (0.2017) & 0.1971 (0.0672) \\
%& & & & n=5 CCA-QA (GTA-99A)
%& 27.33 (18.78) & 59.09 (30.13) & 37.86 (25.20) & 8.18 (5.88) & 100.00 (0.00) & 0.6069 (0.1892) & 0.2030 (0.0833) \\
& \((Q, \bm{A})_{\text{gt}}\) \hfill n=5
& 10.49 (6.31) & 17.08 (13.61) & 28.42 (17.82) & 21.25 (8.64) & 100.00 (0.00) & 0.5656 (0.1375) & 0.0287 (0.0100) \\
& & \((Q, \bm{A})_{\text{max}}\) \hfill n=5
& 9.47 (8.99) & 20.73 (24.04) & 19.14 (17.22) & 14.94 (8.64) & 20.78 (40.59) & 0.7673 (0.0605) & 0.0336 (0.0121) \\
& & \((\gtA,\bm{A})\) \hfill n=5
& 15.34 (14.45) & 89.81 (23.18) & 18.42 (19.76) & 2.74 (4.20) & 100.00 (0.00) & 0.9342 (0.1189) & 0.1369 (0.0672) \\
& & \((\gtA,\tilde{\bm{A}})\) \hfill n=4
& 27.74 (19.02) & 54.15 (30.33) & 42.51 (26.81) & 11.08 (8.49) & 100.00 (0.00) & 0.5290 (0.1968) & 0.2123 (0.0721) \\
& & \((\gtA,\tilde{\bm{A}})\) \hfill n=5
& 25.59 (17.69) & 59.20 (30.71) & 35.74 (24.47) & 7.91 (6.18) & 100.00 (0.00) & 0.5874 (0.1892) & 0.2188 (0.0877) \\
\bottomrule
\end{tabular}}
\vspace*{-1ex}
{\caption
{Intersection verification of semi-supervised methods for automatic reference set construction on the human-annotated validation subset~\(\mathcal{H}_v\), against~\(H\). Values in parentheses denote standard deviation across the set. For \(G\), \(n\) is the number of clusters specified for the agglomerative clustering.}
\label{tab:app-clustering}}
\vspace*{1.5em}
\end{table*}

% \vspace*{-1em}
\paragraph{Using \gls{AMT}}
We show the \gls{AMT} interface and examples of automatic reference sets with their corresponding images and questions in~\cref{fig:amt-interface}.
Each \textsc{hit} consisted of 4 images, each with a question and its corresponding answer set.
Turkers were given \(10\) minutes and \(\$0.10\) to complete each.
Turkers were also required to be based in the \textsc{us}, \textsc{uk}, or Canada (a proxy for English-speaking) and have a \textsc{hit} approval rating of \(\geq 90\%\).
We randomly sample \(1,680\) and \(5,040\) image/question/answer sets from \(\mathcal{H}_v\) and the full validation set, respectively.

\paragraph{Measuring improvement on \acrshort{VD} task}

Our final verification aims to demonstrate the usefulness of the automatic reference sets for learning the \gls{VD} task.
We show that the \gls{CCA}\textsc{-aq} model, trained with the automatic reference sets rather than the human-annotated sets \(H\) achieve equal or better performance on the \gls{VD} task.
We measure performance primarily by the \gls{NDCG} score.
For completeness, we also show performance in the other rank-based metrics.
We extend~\cref{tab:ndcg-cluster} in the main paper by showing results for all automated methods (\(M, \Sigma, G\)) on the validation and test set in~\cref{tab:app-ndcg-cluster} and~\cref{tab:app-ndcg-cluster-val}, respectively.

\end{document}

%% file: maths-basic.tex
\usepackage{scalerel}
\usepackage{mleftright}
\usepackage{dsfont}
\usepackage{bm}

\newcommand{\real}{\mathbb{R}}